\documentclass[a4paper,fleqn]{cas-dc}

\usepackage[numbers]{natbib}

\usepackage{booktabs}
\usepackage{multirow}
\usepackage{colortbl}
\usepackage{xcolor}
\usepackage[utf8]{inputenc}

\usepackage{capt-of}
\usepackage{cuted}

\def\tsc#1{\csdef{#1}{\textsc{\lowercase{#1}}\xspace}}
\tsc{WGM}
\tsc{QE}
\tsc{EP}
\tsc{PMS}
\tsc{BEC}
\tsc{DE}

\begin{document}
\let\WriteBookmarks\relax
\def\floatpagepagefraction{1}
\def\textpagefraction{.001}
\shorttitle{Temporally Consistent Label Interpolation for Robust Surgical Multi-Task Learning under Challenging Conditions}
\shortauthors{G. Kim and J. Park}

\title [mode = title]{Temporally Consistent Label Interpolation for Robust Surgical Multi-Task Learning under Challenging Conditions}                      



\author[1,2]{Garam Kim}
\ead{rka@kist.re.kr}
\credit{Conceptualization, Methodology, Software, Validation, Visualization, Writing - Original draft preparation}

\affiliation[1]{organization={Center for Humanoid Research, Korea Institute of Science and Technology},
                city={Seoul},
                country={Republic of Korea}}

\affiliation[2]{organization={School of Electrical and Electronic Engineering, Yonsei University},
                city={Seoul},
                country={Republic of Korea}}

\author[1]{Juyoun Park}[orcid=0000-0003-3967-9404]
\cormark[1]
\ead{juyounpark@kist.re.kr}
\credit{Conceptualization, Funding acquisition, Project administration, Resources, Supervision, Writing – review and editing}

\cortext[cor1]{Corresponding author}

\nonumnote{This work was supported by the Technology Innovation Program (RS-2024-00443054, Development of a Supermicrosurgical Robot System for Sub-0.8mm Vessel Anastomosis through Human-Robot Autonomous Collaboration in Surgical Workflow Recognition) funded by the Ministry of Trade Industry \& Energy (MOTIE, Korea).}

\begin{abstract}
Effective multi-task learning for surgical scene understanding is fundamentally hindered by annotation granularity mismatch; temporal workflow tasks such as phase recognition, step recognition and anticipation benefit from dense frame-level supervision, whereas pixel-level spatial tasks including instrument segmentation and action recognition are only sparsely annotated on selected keyframes due to prohibitive labeling costs. This supervision imbalance undermines shared representation learning and limits joint optimization across heterogeneous surgical tasks. To address this, we propose Flow-guided Annotation for Robust Operating Scenes (FAROS), 
a flow-guided label interpolation framework, that combines zero-shot segmentation-based mask propagation with optical flow estimation to overcome the limitations of appearance-based propagation under challenging surgical conditions such as occlusion, smoke, and motion blur, generating temporally consistent dense pseudo labels from sparse keyframe annotations. The densified instrument masks and action labels are integrated into a unified Transformer-based multi-task framework that jointly learns surgical phase recognition, step recognition, anticipation, instrument segmentation, and action recognition, enabling balanced optimization between dense temporal supervision and sparse spatial supervision. The label interpolation quality of FAROS is first validated on the DAVIS 2017 benchmark under a sparse ground-truth protocol, confirming robust propagation beyond the surgical domain. Extensive experiments on GraSP, MISAW, and AutoLaparo 
benchmarks further demonstrate that FAROS significantly improves cross-task representation learning and enhances holistic surgical scene understanding performance across spatio-temporal tasks.


\end{abstract}



\begin{keywords}
surgical scene understanding \sep label interpolation \sep multi-task learning \sep surgical image segmentation
\end{keywords}

\maketitle


\section{Introduction}
\label{sec:intro}

Surgical scene understanding has emerged as a fundamental research area in robot-assisted and minimally invasive surgery, aiming to improve surgical safety, procedural efficiency, and intraoperative decision support~\cite{handa2024role, chuchulo2023robotic, morris2005robotic}. Recent advances in computer vision and deep learning have enabled automatic analysis of surgical videos through various tasks, including surgical phase recognition, step recognition, step anticipation, instrument segmentation, and surgical action recognition. These tasks capture complementary aspects of surgical procedures, ranging from high-level workflow semantics to fine-grained instrument interactions, and together provide comprehensive understanding of the intraoperative environment~\cite{maier2017surgical, li2024deep, lalys2014surgical}.

Existing studies have mainly focused on solving these tasks independently. Early workflow analysis methods primarily addressed temporal recognition problems such as surgical phase and step classification using convolutional neural networks (CNNs) and recurrent architectures, while recent Transformer-based approaches further improved long-range temporal reasoning in surgical videos~\cite{gao2021trans, yi2019hard, demir2023deep}. In parallel, encoder–decoder segmentation networks and Transformer-based segmentation models have achieved strong performance in pixel-level surgical instrument localization. 

Although multi-task learning has been increasingly explored in surgical video analysis, unified frameworks that jointly model both dense temporal tasks and sparse pixel-level spatial tasks remain uncommon~\cite{jin2019incorporating, ahmed2024deep}. This is a critical gap, as spatial instrument context and temporal workflow progression are deeply interrelated~\cite{jin2020multi, alabi2025multitask}: instrument presence and motion provide strong cues for phase and step transitions, while procedural context conversely constrains instrument behavior~\cite{twinanda2016endonet, li2024deep, czempiel2020tecno}. Bridging this spatial-temporal divide through joint optimization is therefore essential for holistic surgical scene understanding.

Despite these advantages, effective joint multi-task learning across temporal and spatial tasks is fundamentally hindered by annotation granularity mismatch~\cite{alabi2025multitask}. Workflow-related tasks, such as phase, step, and anticipation recognition, are typically densely annotated along the temporal axis, providing supervision for nearly every video frame~\cite{twinanda2016endonet, ramesh2023weakly}. In contrast, pixel-level spatial tasks including instrument segmentation and action recognition require labor-intensive expert  annotation involving specialized clinical knowledge, making dense pixel-level labeling prohibitively expensive~\cite{antonelli2022medical}. This imbalance between dense temporal supervision and sparse spatial supervision limits effective joint optimization and weakens cross-task representation learning in video-based multi-task frameworks~\cite{nishi2024joint, ruder2017overview}.

A straightforward solution to this problem is to densify sparse spatial annotations through label interpolation~\cite{sestini2023fun, rukundo2024evaluation}. However, directly propagating labels in surgical videos is particularly challenging due to clinically complex intraoperative environments involving occlusion, smoke, motion blur, illumination variation, and frequent instrument appearance and disappearance. Although recent foundation models such as SAM2 demonstrate strong mask propagation capability, their performance often deteriorates under these challenging surgical conditions, leading to temporally inconsistent pseudo labels and unstable spatial supervision~\cite{yin2025memory}.
The core limitation of SAM2 in surgical videos lies in its reliance on appearance-based memory attention, which degrades when instrument appearance changes abruptly due to occlusion, smoke, or specular reflection — conditions that disrupt the visual consistency assumed by the model~\cite{liu2024surgical}. 

\begin{figure*}[t]
    \centering
    \vspace{0.5cm}
    \includegraphics[width=\linewidth]{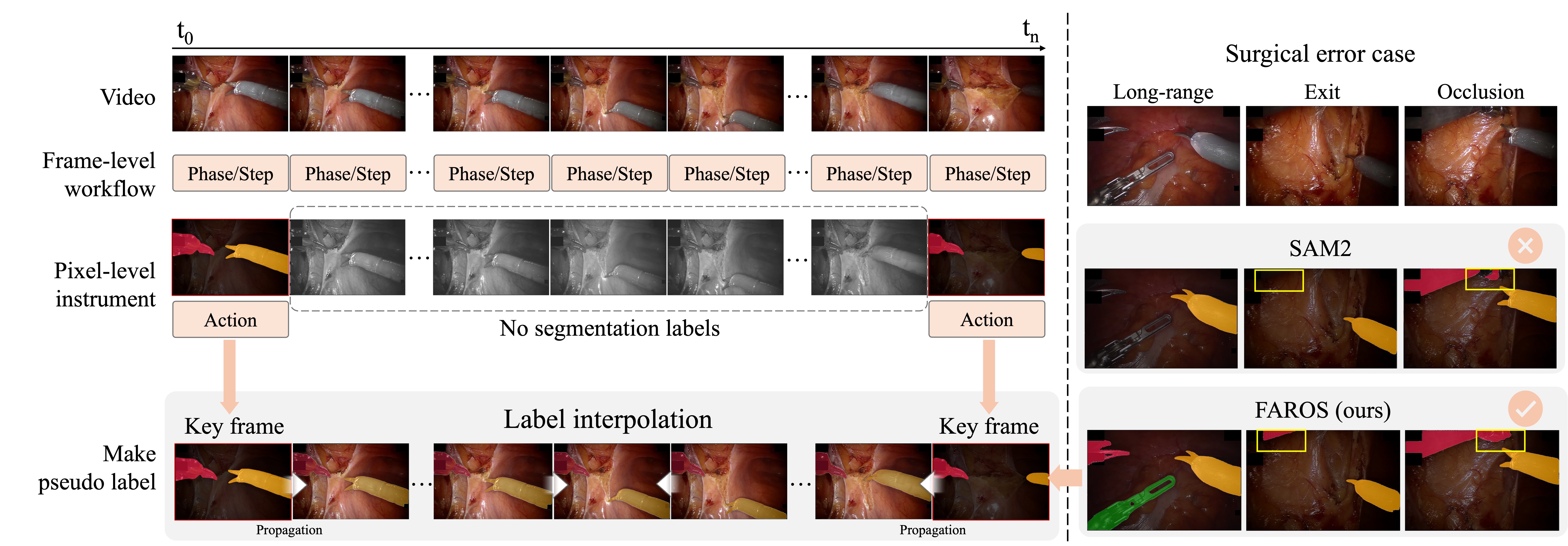}
    \captionof{figure}{Illustration of the annotation granularity mismatch and the motivation for label interpolation. \textbf{(Left)} Frame-level workflow tasks benefit from dense temporal annotations across all frames, whereas pixel-level instrument tasks are sparsely annotated only at selected keyframes. Label interpolation bridges this supervision gap by generating dense pseudo-annotations for all intermediate frames. \textbf{(Right)} Representative failure cases of SAM2-only propagation under challenging surgical conditions (long-range drift, instrument exit, and occlusion), and the corresponding results of FAROS, which successfully recovers accurate masks through flow-guided reprompting.}
    \label{FIG:overview}
\end{figure*}


Optical flow, by contrast, captures explicit geometric motion between frames and remains relatively robust to appearance changes, providing a complementary signal that can detect and correct propagation failures~\cite{teed2020raft, qian2020robust, xie2024appearance}. Prior works have explored optical flow-based label propagation as a means of transferring annotations across frames by exploiting inter-frame motion consistency~\cite{zhao2020learning}. However, flow-based methods alone are susceptible to error accumulation in regions with large appearance changes, occlusion, or non-rigid deformation, which are commonplace in surgical videos~\cite{wu2023accflow, chen2022motion, li2016drift}. Moreover, standalone flow warping lacks semantic awareness, making it difficult to maintain object-level mask coherence under complex instrument interactions. 


To leverage the advantages of both methods while circumventing their inherent drawbacks, we introduce Flow-guided Annotation for Robust Operating Scenes (FAROS), which augments zero-shot segmentation-based mask propagation with optical flow guidance to overcome the limitations of SAM2 alone in clinically complex environments, thereby generating temporally dense and spatially consistent pseudo labels from sparsely annotated surgical frames, as illustrated in Fig.~\ref{FIG:overview}.
Our approach compensates for SAM2's appearance-driven failure modes with geometry-driven spatial priors, thereby generating temporally stable and semantically coherent pseudo labels under challenging surgical conditions.

The interpolated instrument masks and action labels are subsequently integrated into a unified Transformer-based multi-task framework that jointly learns surgical phase recognition, step recognition, anticipation, instrument segmentation, and action recognition.
By bridging the annotation gap between dense temporal supervision and sparse spatial supervision, our framework enables balanced cross-task optimization, where densified spatial priors strengthen spatial grounding for workflow recognition while temporal context regularizes instrument-level predictions, jointly improving performance across surgical tasks under clinically challenging conditions.

Our contributions are summarized as follows:
\begin{enumerate}[\textbullet]
\item We identify annotation granularity mismatch as a fundamental barrier to unified surgical multi-task learning, and propose FAROS, a flow-guided label interpolation framework that generates temporally consistent dense pseudo labels from sparse keyframe annotations under challenging surgical conditions.
\item Building on the densified spatial supervision, we introduce a unified spatio-temporal multi-task learning framework that jointly models workflow-level and region-level tasks within a shared Transformer architecture, enabling spatially grounded cross-task representation learning.
\item We demonstrate on GraSP~\cite{ayobi2024pixel} and MISAW~\cite{huaulme2021micro} benchmarks that FAROS significantly improves cross-task representation learning and enhances holistic surgical scene understanding across all tasks.
\end{enumerate}

\section{Related Work}

\subsection{Surgical Scene Understanding}
Surgical scene understanding has emerged as an important research area with the advancement of robot-assisted and minimally invasive surgery, aiming to improve the safety, efficiency, and precision of surgical procedures~\cite{schreuder2009robotic, maier2017surgical}. By leveraging recent advances in deep learning, surgical scene understanding analyzes surgical videos to interpret intraoperative environments, including surgical instruments, actions, and procedural workflows.
Various tasks have been studied to achieve comprehensive scene understanding, including instrument detection and segmentation, action recognition, and phase and step recognition. Early workflow recognition methods relied on CNNs for frame-level temporal analysis, while recent Transformer-based approaches have improved temporal context modeling~\cite{twinanda2016endonet, gao2021trans, jin2017sv}. For instrument segmentation, encoder--decoder architectures such as U-Net have demonstrated effective pixel-level localization~\cite{unet, shvets2018automatic, laina2017concurrent}.
Despite these advances, most existing approaches address individual tasks independently, limiting their ability to exploit strong correlations among surgical workflow, instrument states, and actions. Recent studies have thus highlighted the need for multi-task frameworks that jointly model complementary spatial and temporal information for holistic surgical scene understanding.

\subsection{Multi-Task Learning}
Multi-task learning (MTL) enhances learning efficiency and predictive performance across multiple tasks compared with training individual models separately~\cite{caruana1997multitask, thrun1995learning}. By enabling cross-task knowledge sharing, representations learned from one task can facilitate the learning of complementary tasks~\cite{zhang2021survey}, and MTL has been widely applied across natural language processing, computer vision, and robotics~\cite{crawshaw2020multi}.
In the medical domain, MTL has been increasingly adopted for surgical scene understanding, where phase, step, instrument, and action recognition exhibit strong inter-task correlations. \cite{valderrama2022towards} introduced the PSI-AVA benchmark and proposed TAPIR, a Transformer-based framework jointly modeling surgical phase, step, instrument detection, and action recognition, demonstrating that shared representations improve holistic scene understanding. However, instrument understanding in TAPIR is limited to bounding box-based detection rather than pixel-level segmentation, constraining the richness of spatial representations available for cross-task learning.
Subsequently, \cite{ayobi2024pixel} introduced the GraSP benchmark and proposed TAPIS, incorporating pixel-level instrument segmentation via localized region embeddings together with global spatio-temporal features. However, spatial and temporal tasks were still learned through separate branches, preventing sufficient interaction between the two scales of representation and limiting bidirectional complementary learning between instrument information and workflow context.

\subsection{Label Interpolation}
Obtaining pixel-level annotations for medical images is highly time-consuming and labor-intensive, often restricting expert annotations to sparsely sampled frames~\cite{antonelli2022medical, zhang2024nasalseg, luo2022semi}. As a result, a large portion of medical video remains unlabeled, creating a strong demand for label interpolation techniques that can effectively exploit unlabeled frames.
Building on this paradigm, conventional label interpolation approaches have primarily relied on self-training, in which pseudo-labels are iteratively generated for unlabeled frames and used to supervise subsequent model updates alongside ground-truth annotations~\cite{tarvainen2017mean, bai2017semi}. 
Beyond iterative pseudo-label generation, another line of research enhances model representation learning through unsupervised regularization on unlabeled data. Consistency-based approaches enforce prediction agreement under various perturbations, including self-ensembling across training iterations, and data recombination augmentations. Nevertheless, semi-supervised segmentation methods still require a labeled subset to achieve acceptable performance, and their propagation strategies remain largely agnostic to inter-frame temporal structure. More recently, the emergence of foundation models has opened new opportunities for annotation-efficient learning, where prompt-driven segmentation models such as SAM and SAM2 can function as powerful pseudo-label generators owing to their strong zero-shot generalization capabilities~\cite{sam, ravi2025sam, zhang2025semisam+}.

Despite these advances, directly applying pseudo-label propagation to surgical videos remains challenging due to the unique visual complexity of intraoperative environments~\cite{yin2025memory}. Several surgical-specific methods have been proposed to address this; \cite{zhao2020learning} introduced an optical flow-based semi-supervised framework for instrument segmentation, but the approach is heavily dependent on accurate flow estimation and degrades under fast instrument motion or large deformations. \cite{sestini2023fun} proposed FUN-SIS, leveraging implicit motion cues and instrument shape priors without spatial annotations, but the method is limited to binary single-instrument segmentation and cannot generalize to multi-class settings. \cite{wei2025segmatch} presented SegMatch, combining consistency regularization with adversarial augmentation, yet it operates on individual frames without explicit temporal propagation. Beyond these method-level limitations, surgical scenes are further characterized by occlusion, motion blur, smoke artifacts, specular reflections, and frequent instrument exit, all of which can degrade propagation quality and introduce temporal inconsistency in pseudo-labels. Furthermore, most existing label propagation methods are designed exclusively for single-task segmentation, with little consideration of how interpolated spatial annotations might facilitate joint optimization across multiple interrelated surgical scene understanding tasks~\cite{alabi2025multitask, zhao2020learning}.

These limitations motivate the development of interpolation strategies capable of generating temporally dense and spatially consistent pseudo-annotations while maintaining motion continuity across complex surgical scenes. 

\section{Methodology}

\subsection{Preliminaries}

\begin{figure*}[t]
    \centering
    \vspace{0.5cm}
    \includegraphics[width=\linewidth]{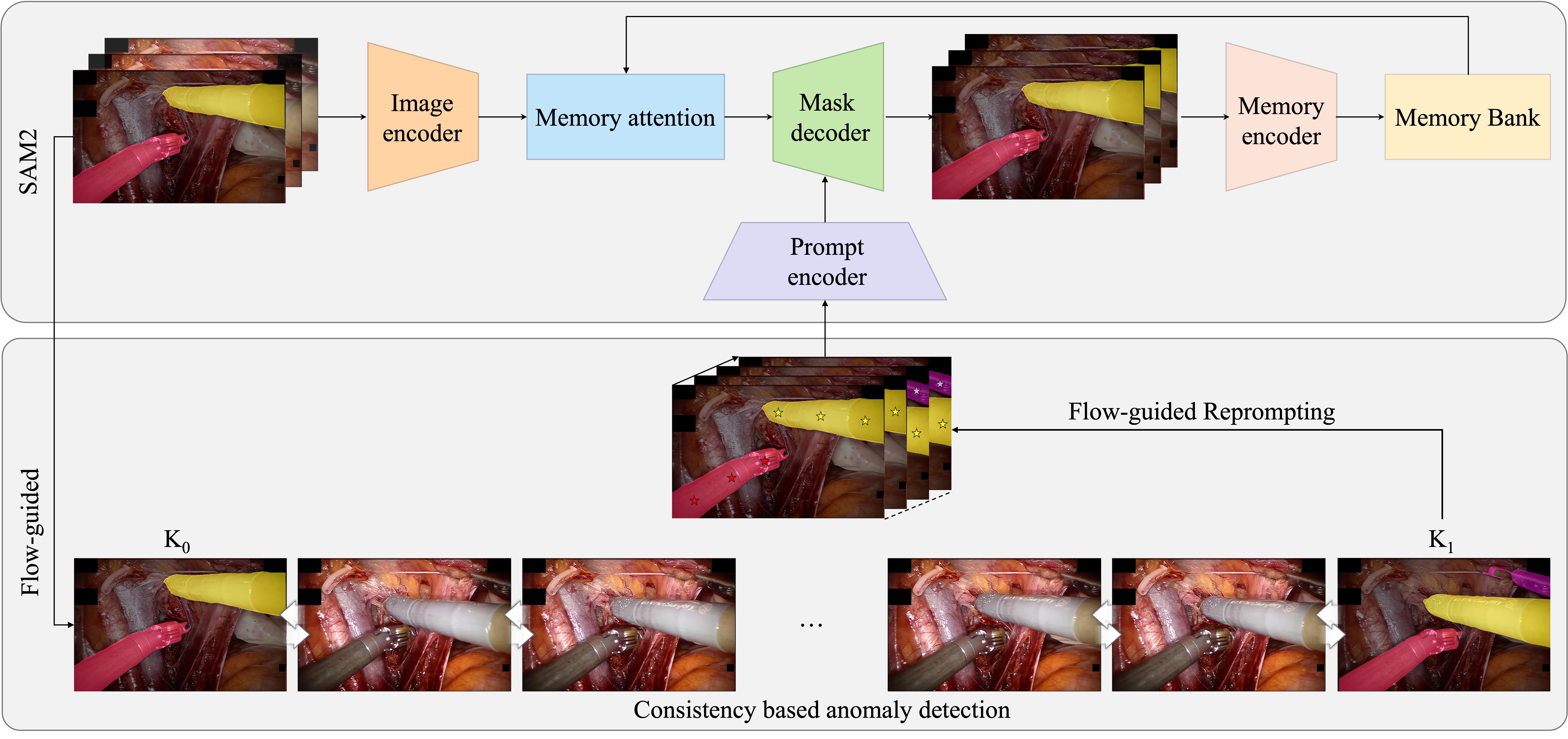}
    \captionof{figure}{Overview of the proposed FAROS pipeline. \textbf{(Top)} The SAM2-based propagation module processes input frames through an image encoder, memory attention, and mask decoder, with a prompt encoder providing corrective spatial prompts. Propagated mask features are stored in a memory bank via the memory encoder for temporally coherent propagation. \textbf{(Bottom)} The flow-guided module operates between keyframes $K_0$ and $K_1$, performing consistency-based anomaly detection across intermediate frames to identify propagation failures. Upon anomaly confirmation, flow-guided reprompting warps the ground-truth mask from the nearest keyframe to generate geometrically grounded corrective prompts, which are fed back into the SAM2 prompt encoder to re-initialize propagation.}
    \label{FIG:fg_seg_pipeline}
\end{figure*}

\noindent\textbf{Semi-Supervised Learning.} 
Semi-supervised learning (SSL) aims to improve model performance under limited annotation budgets by jointly leveraging labeled and unlabeled data~\cite{lee2013pseudo,tarvainen2017mean}. In the context of surgical video understanding, obtaining dense pixel-level annotations is prohibitively expensive, as it requires extensive expert knowledge and manual effort for every frame~\cite{shi2021semi}. SSL addresses this challenge by exploiting the abundant unlabeled frames alongside sparse keyframe annotations, enabling more effective use of available data~\cite{yu2022pseudo, zhao2020learning}. In our framework, we adopt a self-training paradigm in which pseudo labels generated for unlabeled intermediate frames are subsequently used as additional supervision to complement sparse ground-truth annotations.

\noindent\textbf{Optical Flow Estimation.}
Recurrent All-Pairs Field Transforms (RAFT)~\cite{teed2020raft} is an optical flow estimation method that computes dense per-pixel displacement fields between consecutive video frames. RAFT constructs a 4D correlation volume over all pairs of feature vectors and iteratively updates a flow field through a recurrent lookup operator, achieving robust and accurate motion estimation even under challenging conditions. In our framework, we exploit the forward-backward consistency property of RAFT-estimated flows to detect temporally inconsistent mask propagation and to provide geometrically grounded spatial priors for flow-guided reprompting~\cite{gmflow, ren2020unsupervised}.

\noindent\textbf{Segment Anything Model 2 (SAM2).}
SAM2~\cite{ravi2025sam} is a promptable foundation model designed for object segmentation in both images and videos. Built upon a hierarchical image encoder and a memory-augmented decoder, SAM2 maintains a memory bank of past frame features and mask predictions, enabling temporally coherent mask propagation across video frames given sparse prompts such as points, boxes, or masks. Its strong zero-shot generalization capability makes SAM2 particularly attractive as a pseudo-label generator for unlabeled surgical frames~\cite{yu2025sam}. However, as discussed in Section~\ref{sec:intro}, SAM2 alone is susceptible to failure under the challenging visual conditions characteristic of surgical videos, motivating our flow-guided augmentation strategy.

\subsection{FAROS}

We propose a pipeline that generates dense pseudo segmentation labels for surgical videos from sparse keyframe annotations. An overview of the FAROS pipeline is provided in Fig.~\ref{FIG:fg_seg_pipeline}. Given a video with ground-truth masks only at keyframes $\{k_0, k_1, k_2, \ldots\}$, our goal is to produce pseudo labels $\hat{M}_t^c$ for all intermediate non-keyframes $t \in (k_i, k_{i+1})$ where no ground-truth annotation exists, and for all object categories $c \in \mathcal{C}$.

\begin{figure*}[t]
    \centering
    \vspace{0.5cm}
    \includegraphics[width=\linewidth]{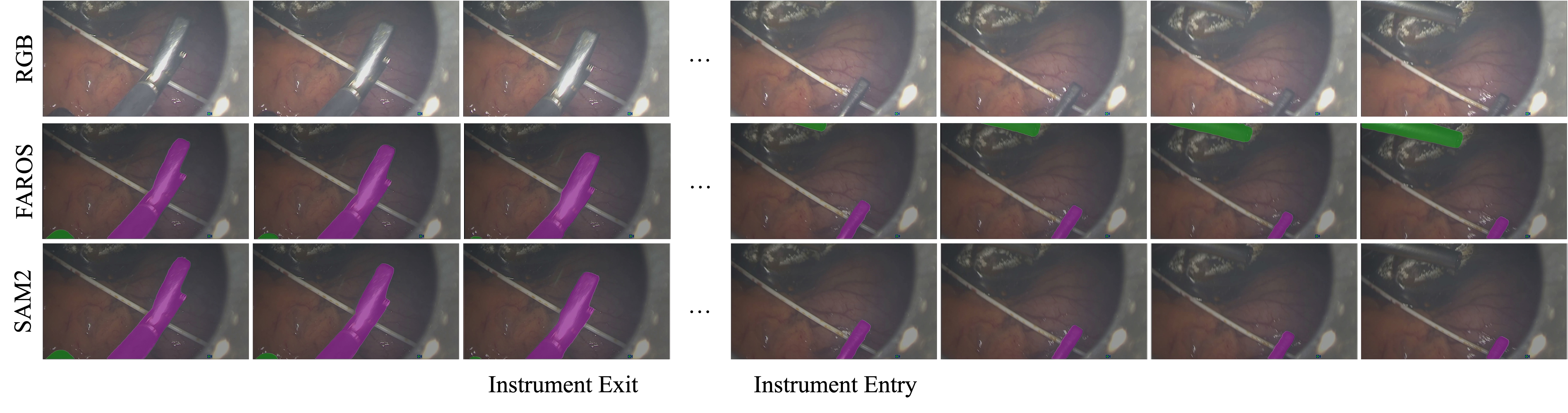}
    \captionof{figure}{Qualitative comparison of mask propagation on an instrument exit case. SAM2-only propagation fails to track instrument disappearance and re-entry, producing temporally inconsistent masks. FAROS detects the propagation failure via flow-guided consistency checking and recovers accurate segmentation through reprompting.}
    \label{FIG:case_exit}
    \vspace{0.5cm}
    \includegraphics[width=\linewidth]{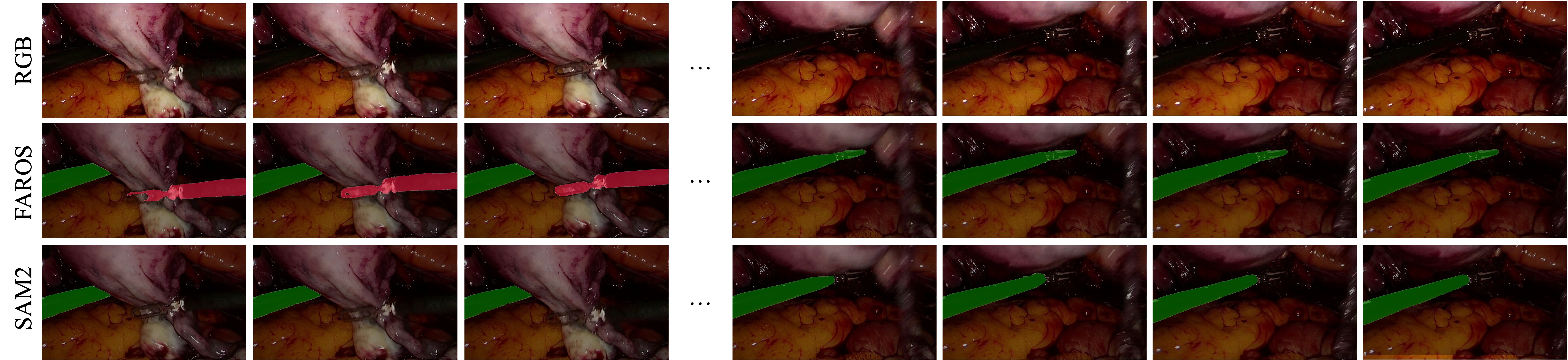}
    \captionof{figure}{Qualitative comparison of mask propagation under illumination variation and blood-induced occlusion. SAM2-only propagation degrades under these appearance disruptions, whereas FAROS maintains robust segmentation by leveraging geometric motion priors to compensate for appearance-driven memory attention failures.}
    \label{FIG:case_occlusion}
\end{figure*}

\paragraph{Bidirectional SAM2 Prompting.}
We process each keyframe segment $(k_0, k_1)$ independently. Naively initializing SAM2 with only the left keyframe $k_0$ causes mask quality to degrade as $t$ approaches $k_1$, since the model relies solely on one-directional temporal context. To address this, we register ground-truth mask prompts from both boundary keyframes into SAM2's memory bank before propagation:
\begin{equation}
    \forall c \in \mathcal{C}: \quad (k_0,\ M_{k_0}^c), \quad (k_1,\ M_{k_1}^c)
\end{equation}
SAM2 then propagates forward from $k_0$ while its memory attention can reference both keyframe anchors, yielding more temporally stable predictions across the full segment.

\paragraph{Flow Consistency-Based Anomaly Detection.}
Due to the inherent characteristics of surgical videos --- including occlusion caused by blood and smoke, as well as abrupt instrument disappearance --- relying solely on SAM2 for mask propagation can lead to failures in these challenging scenarios. To detect such failures, we exploit the forward-backward consistency property of optical flow~\cite{baghbaderani2024temporally, gmflow, ren2020unsupervised}. At each frame $t$, we compute bidirectional flows using RAFT and measure the per-pixel consistency error:
\begin{equation}
    e(\mathbf{p}) = \left\| \mathbf{F}_{t-1 \to t}(\mathbf{p}) + \mathbf{F}_{t \to t-1}\!\left(\mathbf{p} + \mathbf{F}_{t-1 \to t}(\mathbf{p})\right) \right\|_2
\end{equation}
When motion is consistent, a forward-displaced pixel should return to its origin under the backward flow, yielding $e(\mathbf{p})\approx0$. This property breaks in regions undergoing occlusion or large appearance change. The per-object anomaly score is defined as the mean error over the SAM2-predicted mask region:
\begin{equation}
    s_t^c = \frac{1}{|\hat{M}_t^c|} \sum_{\mathbf{p} \in \hat{M}_t^c} e(\mathbf{p})
\end{equation}
An object is flagged as anomalous when $s_t^c > \tau$ persists for $n_\text{confirm}$ consecutive frames, avoiding false positives from isolated noise or transient lighting changes.

\paragraph{Flow-Guided Reprompting.}
Upon anomaly confirmation, we compute the backward flow from the right keyframe $k_1$ to the current frame $t$, and warp the ground-truth mask $M_{k_1}^c$ accordingly:
\begin{equation}
    \tilde{M}_t^c = \text{Warp}(M_{k_1}^c,\ \mathbf{F}_{k_1 \to t})
\end{equation}
Since $M_{k_1}^c$ is a verified ground-truth annotation, $\tilde{M}_t^c$ provides a reliable spatial prior for SAM2 re-initialization at frame $t$, with corrective prompts sampled via distance-transform-based farthest-point sampling. The flow $\mathbf{F}_{k_1 \to t}$ is computed only on anomaly detection to minimize overhead. The final pseudo label defaults to the SAM2 output, replaced by $\tilde{M}_t^c$ when reprompting occurs and the warped mask exceeds the minimum area threshold.

A qualitative comparison of mask propagation between SAM2-only and FAROS is presented in Figs.~\ref{FIG:case_exit} and~\ref{FIG:case_occlusion}. Fig.~\ref{FIG:case_exit} illustrates a challenging instrument exit case, where SAM2-only propagation fails to accurately track instrument disappearance and re-entry, resulting in temporally inconsistent masks. In contrast, FAROS successfully detects the propagation failure through flow-guided consistency checking and recovers accurate segmentation via reprompting. Fig.~\ref{FIG:case_occlusion} demonstrates a case involving illumination variation and blood-induced occlusion, which are characteristic challenges of intraoperative surgical environments. Under these conditions, SAM2-only propagation produces degraded and temporally inconsistent masks, whereas FAROS maintains robust segmentation by leveraging geometric motion priors to compensate for the appearance-driven failure of the memory attention mechanism.

\begin{figure*}[t]
    \centering
    \vspace{0cm}
    \includegraphics[width=0.8\linewidth]{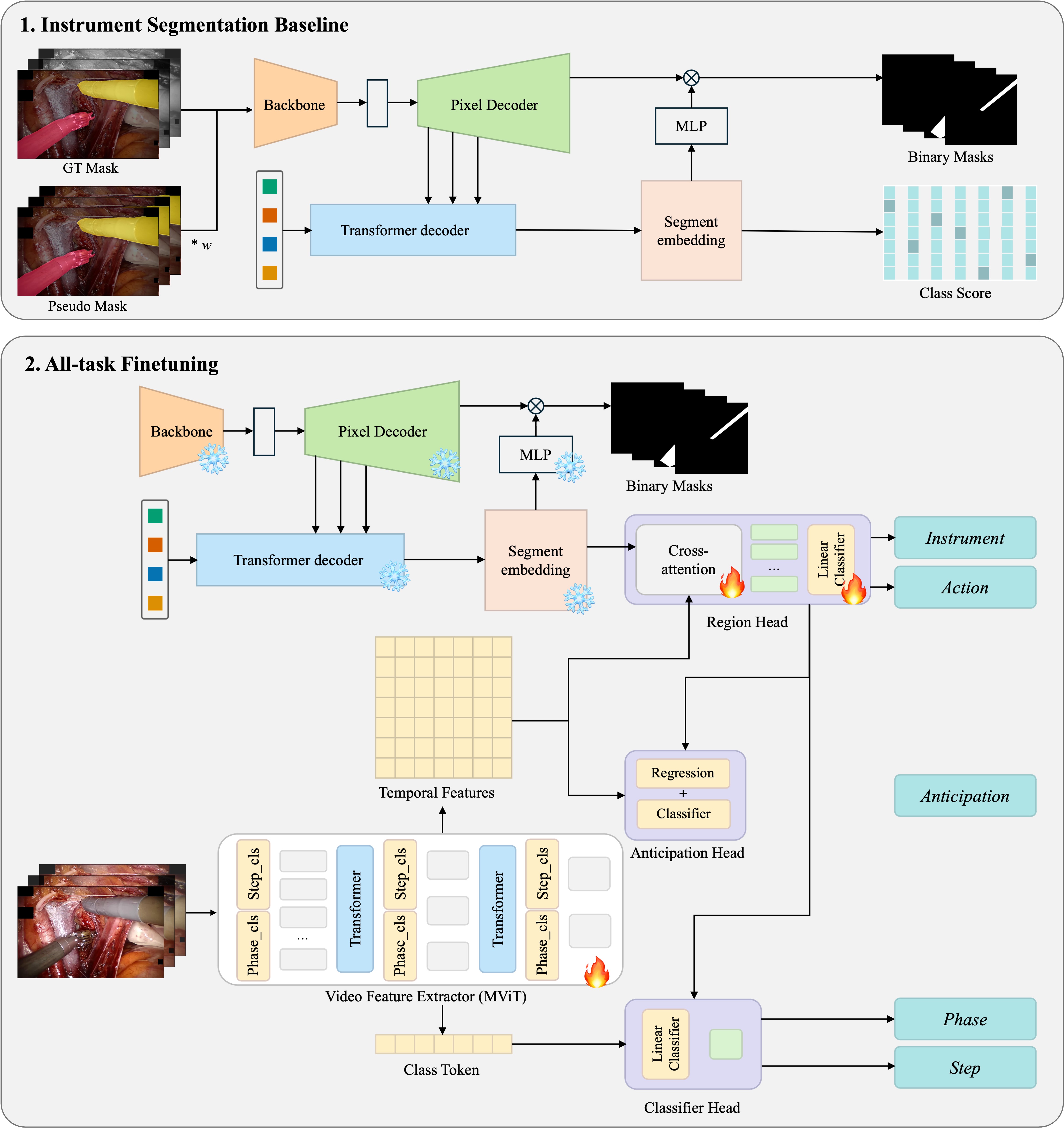}
    \captionof{figure}{Overview of the proposed multi-task learning framework. \textbf{(1. Instrument Segmentation Baseline)} A Mask2Former-based RPN is trained on ground-truth and pseudo masks (weighted by $w$) to produce segment embeddings for mask prediction and class scoring. \textbf{(2. All-task Finetuning)} The frozen RPN provides precomputed region embeddings. An MViT backbone with task-dedicated CLS tokens extracts spatio-temporal features for phase and step recognition, which are combined with region embeddings via cross-attention for instrument and action recognition, and routed to the anticipation head. Flame and snowflake icons denote trainable and frozen modules.}
    \label{FIG:mtl_pipeline}
\end{figure*}

\subsection{MTL for surgical scene understanding}


The dense pseudo-annotations generated by FAROS resolve the annotation sparsity that previously prevented effective joint optimization of temporal workflow tasks and pixel-level spatial tasks. In surgical videos, instrument spatial configurations are tightly coupled to procedural workflow context: the presence and 
arrangement of specific instruments serve as discriminative priors for phase and step recognition, while workflow-level context conversely constrains instrument behavior. By densifying spatial supervision through FAROS, the shared backbone receives consistent instrument-level cues throughout training, enabling the spatial--temporal coupling that was previously disrupted by annotation imbalance. Building on this foundation, we propose a unified multi-task framework that jointly models frame-level surgical workflow understanding and region-level instrument understanding within a single video Transformer architecture, as illustrated in Fig.~\ref{FIG:mtl_pipeline}. Since workflow recognition tasks (phase, step, and anticipation) and region-level tasks (instrument segmentation and action recognition) describe the same surgical events from complementary temporal and spatial perspectives, our framework explicitly exploits their mutual dependencies through three cross-task coupling mechanisms.

\paragraph{Stage 1: Region Proposal Network.}
For the instrument segmentation branch, we employ Mask2Former as the region proposal baseline~\cite{mask2former}. Mask2Former utilizes a Transformer decoder to perform cross-attention between $N$ object queries and image features extracted from the backbone, transforming the queries into $N$ segment-level embeddings. These embeddings provide spatial priors that encode instrument identity as well as local anatomical context. During Stage 2, the Region Proposal Network (RPN) is frozen, and the precomputed region embeddings are reused to enable efficient multi-task training.

\paragraph{Stage 2: Joint Spatio-temporal All-tasks Fine-tuning.}

For each temporal window, the MViT backbone extracts spatio-temporal tokens 
$\mathbf{Z}_t^{\mathrm{ST}}$~\cite{mvit}. Rather than sharing a single CLS token across all workflow tasks, we assign dedicated CLS tokens to each frame-level task:
\begin{equation}
    \mathbf{z}_t^{\mathrm{ph}}, \mathbf{z}_t^{\mathrm{st}} = \mathrm{MViT}(\mathbf{X}_t;\, \theta_{\mathrm{cls}}^{\mathrm{ph}},\, \theta_{\mathrm{cls}}^{\mathrm{st}})
\end{equation}


Each CLS token attends to the shared visual patch tokens throughout all attention blocks of the backbone, allowing task-specific gradients to flow simultaneously and preserving features relevant to both phase and step recognition. Frame-level heads predict surgical phase and step from their respective CLS tokens, while region-level heads combine region embeddings $\mathbf{f}_t^{(i)}$ with $\mathbf{Z}_t^{\mathrm{ST}}$ through cross-attention for instrument classification and action recognition.

\paragraph{Cross-task Conditioning.}
To further exploit inter-task dependencies, we introduce lightweight conditioning modules, all initialized with zero-weight projections to prevent negative transfer during early training.

\noindent{(i) Phase$\rightarrow$Step conditioning.}
Since surgical phases provide high-level procedural context that constrains step-level sub-unit recognition, phase features are injected into the step prediction pathway:
\begin{equation}
    \mathbf{z}_t^{\mathrm{st}} \leftarrow \mathbf{z}_t^{\mathrm{st}} + W_{\mathrm{ph \to st}}\,\mathrm{sg}(\mathbf{z}_t^{\mathrm{ph}})
\end{equation}
where $\mathrm{sg}(\cdot)$ denotes stop-gradient, ensuring that step loss does not back-propagate through the phase head. This allows phase representations to inform step prediction only after the phase head has sufficiently converged.

\noindent{(ii) Spatial$\rightarrow$Temporal conditioning.}
Instrument presence and spatial configuration provide strong cues for both phase and step recognition. Region embeddings are pooled across valid instrument instances and injected into both workflow heads:
\begin{equation}
    \mathbf{c}_t^{\mathrm{spatial}} = \frac{1}{|\mathcal{V}|}\sum_{i \in \mathcal{V}} \mathbf{f}_t^{(i)}
\end{equation}
\begin{equation}
    \mathbf{z}_t^{\mathrm{ph}} \leftarrow \mathbf{z}_t^{\mathrm{ph}} + W_{\mathrm{sp \to ph}}\,\mathbf{c}_t^{\mathrm{spatial}}, \quad
    \mathbf{z}_t^{\mathrm{st}} \leftarrow \mathbf{z}_t^{\mathrm{st}} + W_{\mathrm{sp \to st}}\,\mathbf{c}_t^{\mathrm{spatial}}
\end{equation}

where $\mathcal{V}$ denotes the set of valid instrument instances. This mechanism grounds temporal workflow predictions in pixel-level instrument understanding, enabling the framework to leverage the dense spatial supervision provided by the interpolated pseudo-labels.

The overall objective is defined as the weighted sum of all task-specific losses:

\begin{equation}
\mathcal{L} =
\lambda_{\mathrm{ph}}\mathcal{L}_{\mathrm{ph}}
+\lambda_{\mathrm{st}}\mathcal{L}_{\mathrm{st}}
+\lambda_{\mathrm{anti}}\mathcal{L}_{\mathrm{anti}}
+\lambda_{\mathrm{seg}}\mathcal{L}_{\mathrm{seg}}
+\lambda_{\mathrm{act}}\mathcal{L}_{\mathrm{act}}
\end{equation}

Each loss is activated only when the corresponding annotations are available. This design enables balanced optimization across heterogeneous temporal and spatial supervision signals.

\begin{table}[t]
\centering
\caption{Comparison of datasets. Temporal Frames denotes the number of frames annotated for frame-level tasks (phase and step recognition), and Spatial Frames denotes the number of frames annotated for pixel-level tasks (instrument segmentation and action recognition). Recog. denotes recognition.}
\small
\renewcommand{\arraystretch}{0.8}
\resizebox{\columnwidth}{!}{%
\begin{tabular}{lcccccc}
\toprule
\midrule
Dataset
 & \begin{tabular}[c]{@{}c@{}}Phase \\ Recog.\end{tabular}
 & \begin{tabular}[c]{@{}c@{}}Step \\ Recog.\end{tabular}
 & \begin{tabular}[c]{@{}c@{}}Action \\ Recog.\end{tabular}
 & \begin{tabular}[c]{@{}c@{}}Instrument \\ Segmentation\end{tabular}
 & \begin{tabular}[c]{@{}c@{}}Temporal \\ Frames\end{tabular}
 & \begin{tabular}[c]{@{}c@{}}Spatial \\ Frames\end{tabular} \\
\midrule
GraSP \cite{ayobi2024pixel}                                   & $\checkmark$ & $\checkmark$ & $\checkmark$ & $\checkmark$ & 116,515  & 3,449  \\
MISAW \cite{huaulme2021micro, misaw_segdata} & $\checkmark$ & $\checkmark$ & $\checkmark$ & $\checkmark$ & 164,275  & 2,999  \\
AutoLaparo \cite{wang2022autolaparo}                          & $\checkmark$ &              &              & $\checkmark$ & 83,243   & 1,800  \\
\bottomrule
\end{tabular}%
}
\label{tab:dataset_comparison}
\end{table}

\section{Experiments}

\subsection{Datasets}

We evaluate our framework on multiple surgical video datasets covering diverse task annotations and surgical domains, as summarized in Table~\ref{tab:dataset_comparison}.

For our primary evaluation, we adopt GraSP~\cite{ayobi2024pixel} and MISAW~\cite{huaulme2021micro} as the main benchmarks, as both datasets provide comprehensive multi-level annotations covering frame-level workflow tasks (phase, step, and anticipation) alongside pixel-level spatial tasks (instrument segmentation, action recognition). This multi-level annotation coverage enables rigorous evaluation of holistic surgical scene understanding across complementary temporal and spatial scales. GraSP covers the domain of robot-assisted surgery on human subjects, while MISAW spans robot-assisted microsurgical training scenarios, together providing diverse evaluation conditions for our framework. Note that while MISAW provides workflow annotations for surgical phases, steps, and actions, it lacks pixel-level spatial annotations in its original release. To address this, we additionally constructed instance segmentation annotations of surgical instruments for the MISAW dataset, and use this extended version throughout our experiments~\cite{misaw_segdata}. 

AutoLaparo~\cite{wang2022autolaparo} provides both phase recognition and instrument segmentation annotations; however, these two annotation types are provided independently without co-annotation at the same frames, precluding direct joint multi-task evaluation. We therefore utilize AutoLaparo as an auxiliary benchmark to assess the downstream segmentation performance of our flow-guided label interpolation pipeline in isolation.

Although various surgical datasets exist, many are limited in their annotation scope. For example, Cholec80~\cite{twinanda2016endonet} provides only phase-level annotations in the laparoscopic domain, limiting its applicability to workflow-level supervision alone. Similarly, EndoVis2018~\cite{allan20202018} offers pixel-level instrument segmentation and instrument-centric action recognition annotations, covering short-term spatial tasks but lacking any frame-level workflow supervision. Since these existing datasets generally do not provide annotations spanning both temporal workflow and spatial instrument understanding, they are excluded from our primary evaluation.

\subsection{Implementation Details}

All experiments are implemented in PyTorch using an NVIDIA RTX A6000 GPU. 
Our framework consists of two components: (1) FAROS, a flow-guided label interpolation pipeline that generates dense pseudo-annotations from sparse keyframe annotations as a preprocessing step, and (2) a multi-task learning pipeline comprising a Mask2Former region-proposal network that produces instance masks and per-region embeddings, followed by a dual-loader multi-task model that jointly learns long- and short-term tasks on top of those embeddings.

\paragraph{FAROS: Flow-guided Label Interpolation.}
Label interpolation is performed as a preprocessing step prior to Stage~1 training. We use SAM2 (Hiera-B+) as the promptable segmentation model and RAFT (FlyingThings) as the optical flow estimator. The anomaly detection threshold and confirmation window are set to $\tau = 3.0$ and $n_{\mathrm{confirm}} = 3$ across all datasets. Upon anomaly confirmation, 3 positive point prompts are sampled from the warped mask via distance-transform-based farthest-point sampling for flow-guided reprompting. Keyframe intervals are set to 30, 25, and ${\sim}$1{,}000 frames for MISAW, AutoLaparo, and GraSP, respectively, reflecting the varying annotation densities of each dataset.

\paragraph{Stage 1: Mask2Former region proposals.}
We adopt Mask2Former with a Swin-L backbone (window size 12, $384^2$ pretraining, ImageNet-22K) as the region-proposal network \cite{swin}. Training uses AdamW with $\eta_{\mathrm{base}}=1\times 10^{-4}$, weight decay $0.05$, a $0.1$ backbone learning rate (LR) multiplier, gradient clipping at $0.01$. The resulting query embeddings and masks are cached and consumed unchanged by Stage 2.


\paragraph{Pseudo-labeled Stage-1 training.}
Because dense annotations cover only a small fraction of frames, we mix the two with a deliberately conservative pseudo weight. Each batch is drawn from a joint pool in which a pseudo frame is sampled at a rate $w_{\mathrm{pseudo}}$ relative to a ground-truth (GT) frame, so the realized per-batch pseudo share is $\rho = w_{\mathrm{pseudo}}\cdot N_{\mathrm{pseudo}}/(N_{\mathrm{GT}}+w_{\mathrm{pseudo}}\cdot N_{\mathrm{pseudo}})$. We set $w_{\mathrm{pseudo}}=3\times 10^{-3}$ for all three datasets MISAW, GraSP and AutoLaparo, as the pseudo frames substantially outnumber GT frames in all datasets and a conservative weight prevents pseudo-label noise from dominating the training signal. To preserve the effective batch size under the extra forward graph required by pseudo samples, we halve the per-GPU image batch size and accumulate 2--4 gradient steps without an optimizer update, proportionally extending the total number of training iterations so that the total ground-truth throughput matches the GT-only baseline. 

\paragraph{Stage 2: Dual-loader multi-task training.}
While long-term tasks (phase, step, anticipation) are densely labeled at every frame and require clips spanning several seconds, short-term tasks (instrument, action) require short, dense clips centered on annotated keyframes. To efficiently handle both tasks simultaneously, we adopt a dual-loader strategy. The long-term loader samples at rate 30 and activates only the frame-level heads, while the short-term loader samples at rate 1 with dense clips centered on keyframes, loading the cached Stage-1 region embeddings alongside each clip and activating only the region-level heads. Each loader performs an independent forward–backward pass, and gradients from both passes are accumulated into the same parameters before the optimizer step. We apply heterogeneous learning rates across three parameter groups. Using a base LR of $\eta_{\mathrm{base}} = 1.25 \times 10^{-2}$, the video-pretrained backbone is scaled by $5 \times 10^{-3}$, the pretrained long-term heads by $1 \times 10^{-3}$, and the randomly initialized short-term heads by $1.0$. The larger total weight on the long-term side ($0.75$ vs.\ $0.35$) compensates for the weaker per-parameter gradient signal of the long-term heads, which rely solely on the global CLS tokens. This particular combination of weights was selected based on empirical search over multiple configurations.

\subsection{Evaluation Metrics}

For surgical phase and step recognition, we report mean Average Precision (mAP), F1-score, and Accuracy, computed on frames sampled at 1 fps. For instrument segmentation, we adopt both instance-based and semantic segmentation metrics. Instance-level evaluation follows the PASCAL VOC framework~\cite{everingham2015pascal}, reporting instance mAP. Semantic segmentation performance is measured using Mean Intersection over Union (mIoU), Intersection over Union (IoU), and Mean Class Intersection over Union (mcIoU)~\cite{nwoye2020recognition}. For action recognition, we follow the AVA evaluation protocol~\cite{gu2017}, reporting instance-level mean average precision at 0.5 IoU threshold (mAP@0.5 IoU$_{\text{box}}$) applied to instrument bounding boxes, reflecting the instrument-centric nature of surgical actions. For step anticipation, the objective is to predict the remaining time until the next step transition. Following IIA-Net~\cite{rivoir2020rethinking}, we report frame-based MAE:
\begin{align}
    \text{MAE}_{\text{in}} &= \frac{1}{T} \sum_{i}^{T} \text{MAE}(f_i,\ r(\tau(x))), \quad 0 < r(\tau(x)) < h
\end{align}
where $f_i$ denotes the model prediction and $r(\tau(x))$ is the ground-truth remaining time at the current timestamp. $\text{MAE}_{\text{in}}$ measures the mean error over all anticipated frames within the horizon $h$. $h$ restricts evaluation to the earliest $10\%$ of the horizon, capturing performance in the most actionable anticipation window for surgical assistance.

\subsection{FAROS Evaluation on General Video Segmentation}

\begin{figure*}
    \centering
    \includegraphics[width=0.9\linewidth]{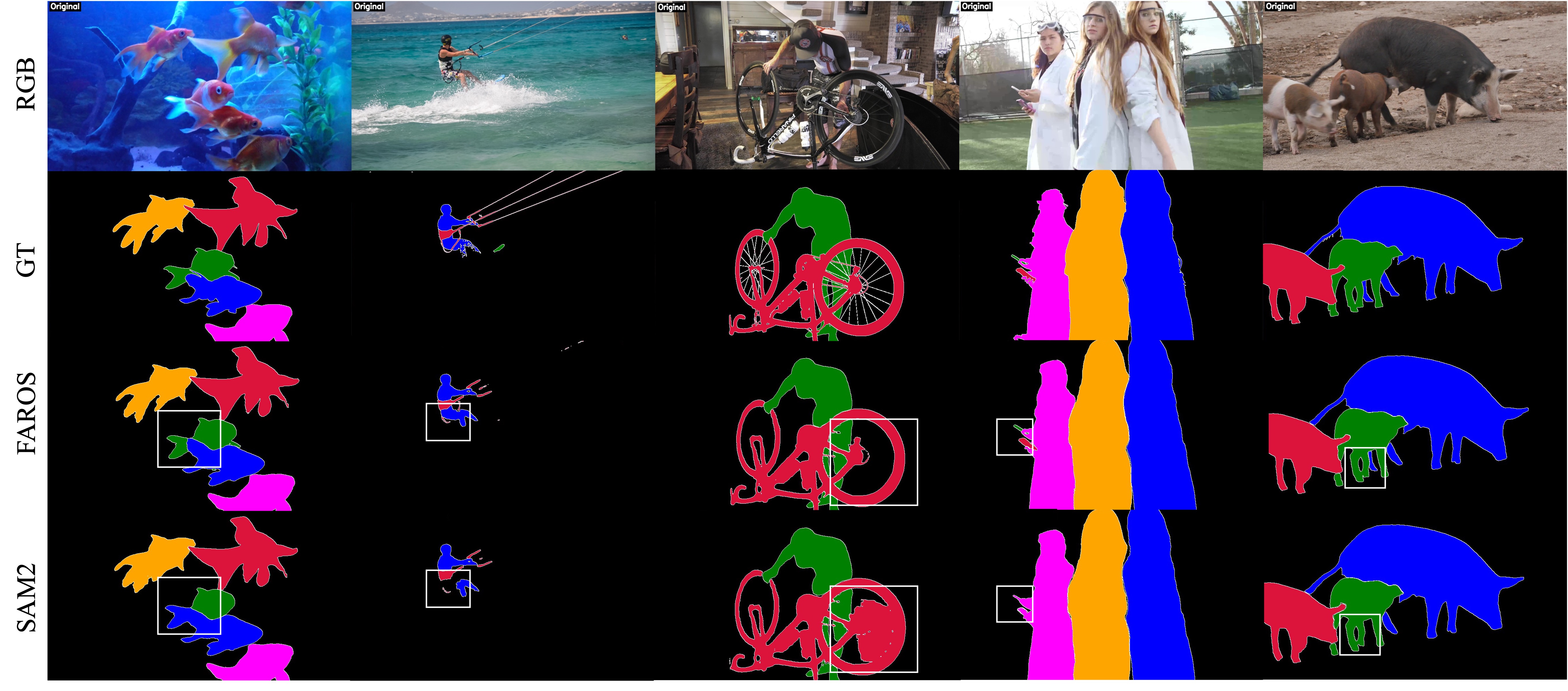}
    \captionof{figure}{Qualitative comparison of mask propagation results on the DAVIS 2017 validation 
    set under the sparse ground-truth interpolation protocol (ground truth provided every 30 
    frames). FAROS maintains temporally consistent segmentation across diverse challenging 
    scenarios including fast-moving objects, partial occlusion, and large inter-frame 
    appearance changes, while SAM2 baseline propagation exhibits mask drift and identity 
    confusion under long-range propagation.} 
    \label{FIG:fg_case_davis}
\end{figure*}

To decouple the evaluation of label interpolation quality from downstream surgical task performance, we assess FAROS on the DAVIS 2017~\cite{pont20172017} validation set, which comprises 30 sequences covering diverse real-world scenes with complex object motion, occlusion, and appearance change. Performance is measured using the standard region similarity $\mathcal{J}$ and contour accuracy $\mathcal{F}$ metrics, along with their mean $\mathcal{J\&F}$.

\paragraph{Video Object Segmentation.}
We first evaluate FAROS under the standard VOS protocol, where only the first-frame annotation is provided and the model propagates masks through the entire sequence. As shown in Table~\ref{tab:davis_standard}, FAROS achieves competitive performance against prior memory-based VOS methods and outperforms the promptable segmentation baseline, confirming that flow-guided augmentation consistently improves propagation quality even under the standard long-range setting.

\begin{table}[t]
\centering
\caption{Comparison on DAVIS 2017 validation set under the 
standard semi-supervised VOS protocol (first-frame annotation only). 
$\mathcal{J}$: region similarity; $\mathcal{F}$: contour accuracy.}
\label{tab:davis_standard}
\footnotesize
\renewcommand{\arraystretch}{1.05}
\begin{tabular}{lccc}
\toprule
Method & $\mathcal{J\&F}$ & $\mathcal{J}$ & $\mathcal{F}$ \\
\midrule
STM~\cite{stm}      & 81.8 & 79.2 & 84.3 \\
HMMN~\cite{HMMN} & 84.7 & 81.9 & 87.5 \\
STCN~\cite{STCN}   & 85.4 & 82.2 & 88.6 \\
RDE~\cite{RDE}        & 84.2 & 80.8 & 87.5 \\
XMem~\cite{Xmem}         & 86.2 & 82.9 & 89.5 \\
ISVOS~\cite{ISVOS}         & 88.2 & 84.5 & 91.9 \\
SAM2~\cite{ravi2025sam}(Hiera-B+)           & 90.2 & 88.4 & 92.4 \\
\midrule
FAROS (Ours)                      & \textbf{90.5} & \textbf{88.5} & \textbf{92.6} \\
\bottomrule
\end{tabular}
\end{table}

\begin{table}[t]
\centering
\caption{Comparison on DAVIS 2017 validation set under the 
sparse-GT interpolation protocol (ground-truth provided every 
30 frames; seeded frames excluded from evaluation). This protocol 
reflects the surgical annotation scenario and is not directly 
comparable to standard semi-supervised VOS methods.
$\mathcal{J}$: region similarity; $\mathcal{F}$: contour accuracy.}
\label{tab:davis_sparse}
\footnotesize
\renewcommand{\arraystretch}{1.05}
\begin{tabular}{lccc}
\toprule
Method & $\mathcal{J\&F}$ & $\mathcal{J}$ & $\mathcal{F}$ \\
\midrule
SAM2 baseline \cite{ravi2025sam}(Hiera-B+) & 91.7 & 89.4 & 93.9 \\
\midrule
FAROS (Ours) & \textbf{93.3} & \textbf{91.1} & \textbf{95.6} \\
\bottomrule
\end{tabular}
\end{table}

\begin{table*}[t]
\centering
\caption{Quantitative results on the GraSP dataset.
Bold denotes the best performance.
$\downarrow$ indicates lower is better.}
\label{tab:grasp}
\resizebox{\textwidth}{!}{%
\begin{tabular}{lccccccccccccc}
\toprule
\multicolumn{14}{c}{\textbf{GraSP}} \\
\midrule
 & \multirow{2}{*}{MTL} & \multirow{2}{*}{Inter}
 & \multicolumn{3}{c}{Phase}
 & \multicolumn{3}{c}{Step}
 & \multicolumn{1}{c}{Anti (h=1)}
 & \multicolumn{3}{c}{Instrument}
 & \multicolumn{1}{c}{Action} \\
\cmidrule(lr){4-6}\cmidrule(lr){7-9}\cmidrule(lr){10-10}\cmidrule(lr){11-13}\cmidrule(lr){14-14}
 & & & mAP & f1 & acc & mAP & f1 & acc &  $\text{MAE}_{in}(\downarrow)$ & mIoU & IoU & mcIoU & mAP \\
\midrule
Single tasks    & $\times$ & $\times$ & 78.211 & \textbf{65.614} & 74.198 & 49.365 & 42.516 & 57.603 & \textbf{0.322} & \textbf{86.553} & 83.418 & \textbf{77.445} & \textbf{25.164} \\
Ph+St      & 2        & $\times$ & 78.317 & 64.735 & 73.634 & 50.031 & 45.710 & \textbf{58.666} & - & - & - & - & - \\
Ph+St+Anti           & 3        & $\times$ & 77.992 & 64.263 & \textbf{74.294} & 51.305 & 45.356 & 59.279 & 0.362 & - & - & - & - \\
All (w/o inter) & 5        & $\times$ & 75.676 & 63.041 & 73.012 & 49.060 & 44.554 & 57.882 & 0.342 & 85.570 & 82.031 & 69.201 & 23.402 \\
All (w/ inter)  & 5        & $\circ$  & \textbf{78.773} & 64.736 & 73.546 & \textbf{51.445} & \textbf{46.046} & 59.060 & 0.398 & 85.227 & \textbf{83.811} & 73.952 & 23.986 \\
\bottomrule
\end{tabular}%
}
\end{table*}

\begin{table*}[t]
\centering
\caption{Quantitative results on the MISAW dataset.
Bold denotes the best performance.
$\downarrow$ indicates lower is better.}
\label{tab:misaw}
\resizebox{\textwidth}{!}{%
\begin{tabular}{lccccccccccccc}
\toprule
\multicolumn{14}{c}{\textbf{MISAW}} \\
\midrule
 & \multirow{2}{*}{MTL} & \multirow{2}{*}{Inter}
 & \multicolumn{3}{c}{Phase}
 & \multicolumn{3}{c}{Step}
 & \multicolumn{1}{c}{Anti (h=1)}
 & \multicolumn{3}{c}{Instrument}
 & \multicolumn{1}{c}{Action} \\
\cmidrule(lr){4-6}\cmidrule(lr){7-9}\cmidrule(lr){10-10}\cmidrule(lr){11-13}\cmidrule(lr){14-14}
 & & & mAP & f1 & acc & mAP & f1 & acc & $\text{MAE}_{in}(\downarrow)$ & mIoU & IoU & mcIoU & mAP \\
\midrule
Single tasks   & $\times$ & $\times$ & 97.851 & 90.797 & 94.653 & 79.186 & 71.312 & 73.338 & 0.284 & 79.094 & 78.745 & 76.209 & 27.961 \\
Ph+St     & 2        & $\times$ & 97.243 & \textbf{91.767} & \textbf{95.605} & 81.451 & 73.157 & 75.739 & - & - & - & - & - \\
Ph+St+Anti          & 3        & $\times$ & 96.672 & 88.337 & 91.643 & 81.500 & 70.970 & 71.414 & 0.390 & - & - & - & - \\
All (w/o inter) & 5       & $\times$ & 95.723 & 88.245 & 90.957 & 80.425 & 71.879 & 74.758 & 0.364 & 75.476 & 74.937 & 71.952 & 26.649 \\
All (w/ inter)  & 5       & $\circ$  & \textbf{98.132} & 91.577 & 94.760 & \textbf{84.022} & \textbf{74.948} & \textbf{77.088} & \textbf{0.204} & \textbf{79.544} & \textbf{79.507} & \textbf{79.039} & \textbf{28.798} \\
\bottomrule
\end{tabular}%
}
\end{table*}

\paragraph{Sparse GT Label Interpolation.}
We further validate FAROS under a sparse ground-truth (GT) interpolation protocol that directly reflects the surgical annotation scenario: ground-truth masks are provided at every 30th frame, and the model interpolates masks for all intermediate frames. Building on the propagation capability validated under the standard VOS protocol, we apply FAROS to this more challenging setting that mirrors the sparse keyframe structure of surgical datasets. We compare FAROS against SAM2 under identical conditions, as standard VOS methods assume single first-frame initialization and are not applicable to this bidirectional keyframe-anchored setting. As shown in Table~\ref{tab:davis_sparse} and Fig.~\ref{FIG:fg_case_davis}, FAROS consistently outperforms the baseline across all metrics, validating that flow-guided anomaly detection and geometry-driven reprompting effectively recover propagation failures introduced by long inter-keyframe intervals.

\paragraph{Clinical Relevance of Challenging-Case Recovery.}

The DAVIS 2017 cases in which FAROS recovers propagation failures—fast motion, partial occlusion, frequent object exit and re-entry, and large inter-frame appearance change—are direct analogues of intraoperative visual disruptions. The instrument exit and re-entry in Fig.~\ref{FIG:case_exit} mirror surgical instruments leaving the endoscopic field of view or moving behind anatomical structures, while the appearance-driven drift in Fig.~\ref{FIG:case_occlusion} corresponds to the illumination variation, specular reflection, and blood- and smoke-induced occlusion common in surgery. In both cases SAM2's appearance-based memory attention degrades when visual consistency breaks, whereas FAROS uses geometric motion priors to detect these failures and re-anchor propagation to verified keyframes. Validating on these general-domain cases under a sparse ground-truth protocol that mirrors the surgical annotation scenario thus confirms that FAROS recovers precisely the failure modes SAM2 cannot handle alone, evidencing its robustness under clinically challenging conditions.

\subsection{Multi-task Learning Results}

We evaluate the proposed framework on GraSP and MISAW as primary benchmarks, and AutoLaparo as an auxiliary benchmark for label interpolation quality. All configurations follow the ablation structure defined in Tables~\ref{tab:grasp} and~\ref{tab:misaw}, progressively adding tasks and enabling interpolation to isolate the contribution of each component.

\begin{figure*}[t]
    \centering
    \vspace{0.5cm}
    \includegraphics[width=0.8\linewidth]{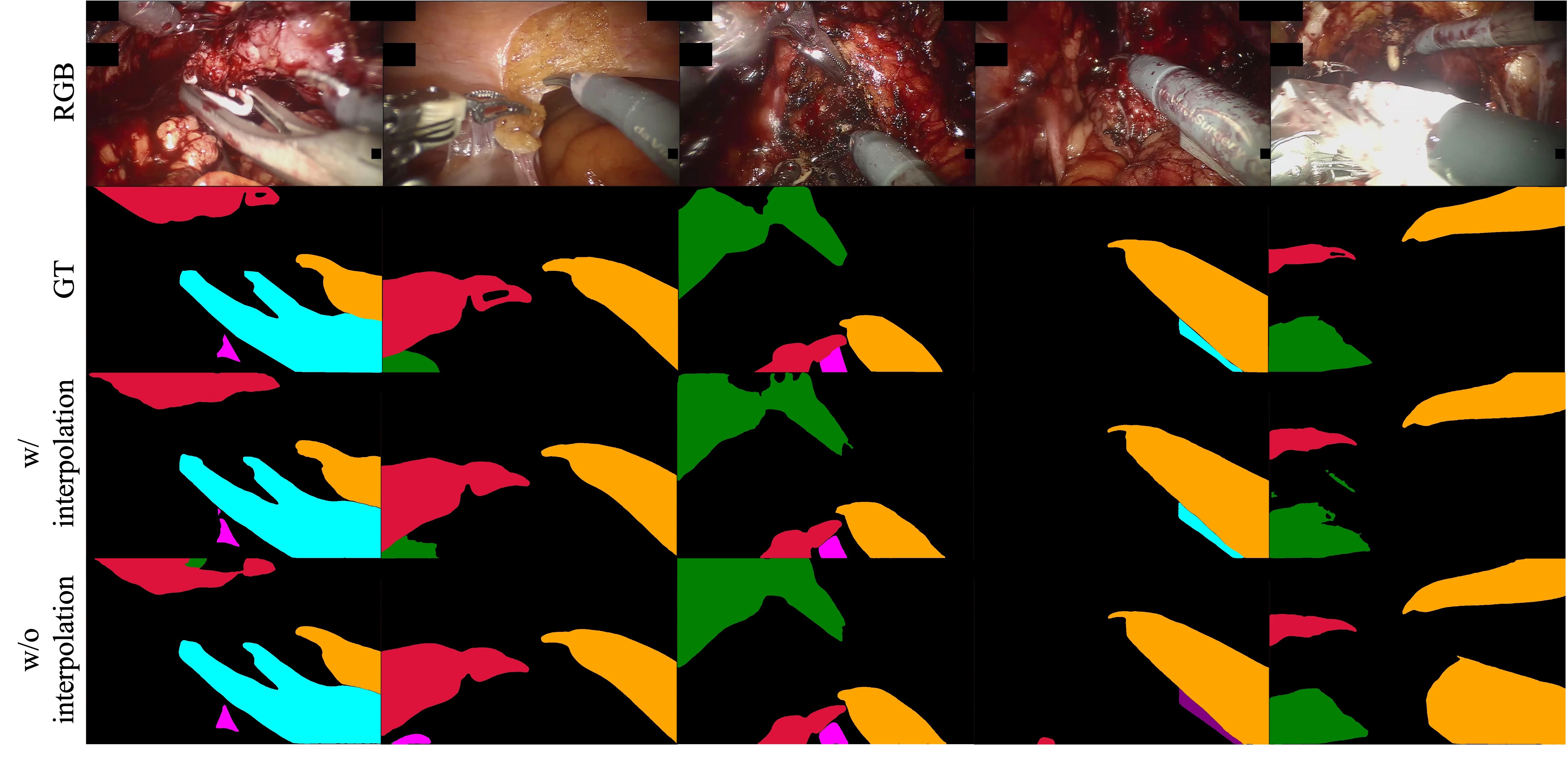}
    \captionof{figure}{Visualization of Instrument Segmentation Results for Comparison on the GraSP Dataset} 
    \label{FIG:grasp_seg}
\end{figure*}

\paragraph{GraSP.}
Table~\ref{tab:grasp} presents quantitative results on GraSP. Multi-task learning without label interpolation (All w/o inter) degrades performance relative to single-task baselines, reflecting the adverse effect of annotation imbalance on joint optimization. Our full model with flow-guided label interpolation (All w/ inter) demonstrates pronounced improvements in workflow-level tasks, with particularly strong gains in step recognition, highlighting that densifying pixel-level spatial annotations provides the shared backbone with richer instrument-level cues across a broader range of procedural contexts. While a marginal mIoU decline is observed compared to All (w/o inter), qualitative analysis in Fig.~\ref{FIG:grasp_seg} reveals that All (w/ inter) produces visually superior segmentation results with more accurate class assignments and fewer false positive masks. We attribute the slight quantitative gap to intermittent pseudo-label imperfections under particularly challenging surgical conditions, which are isolated and do not reflect the overall quality of the generated pseudo-labels. The supplementary video further demonstrates this qualitative advantage, showing temporally consistent mask predictions and robust instrument detection and segmentation even under smoke condition.

\begin{table}[h]
\centering
\caption{Instrument segmentation performance on AutoLaparo. Bold denotes the best performance.}
\label{tab:autolaparo}
\footnotesize
\renewcommand{\arraystretch}{1.0}
\begin{tabular}{lccc}
\toprule
\multicolumn{4}{c}{\textbf{AutoLaparo}} \\
\midrule
 & IoU & mIoU & mcIoU \\
\midrule
w/o interpolation & 83.546          & 86.029          & \textbf{81.963} \\
w/ interpolation  & \textbf{85.516} & \textbf{87.544} & 81.787 \\
\bottomrule
\end{tabular}
\end{table}

\begin{figure*}
    \centering
    \includegraphics[width=0.75\linewidth]{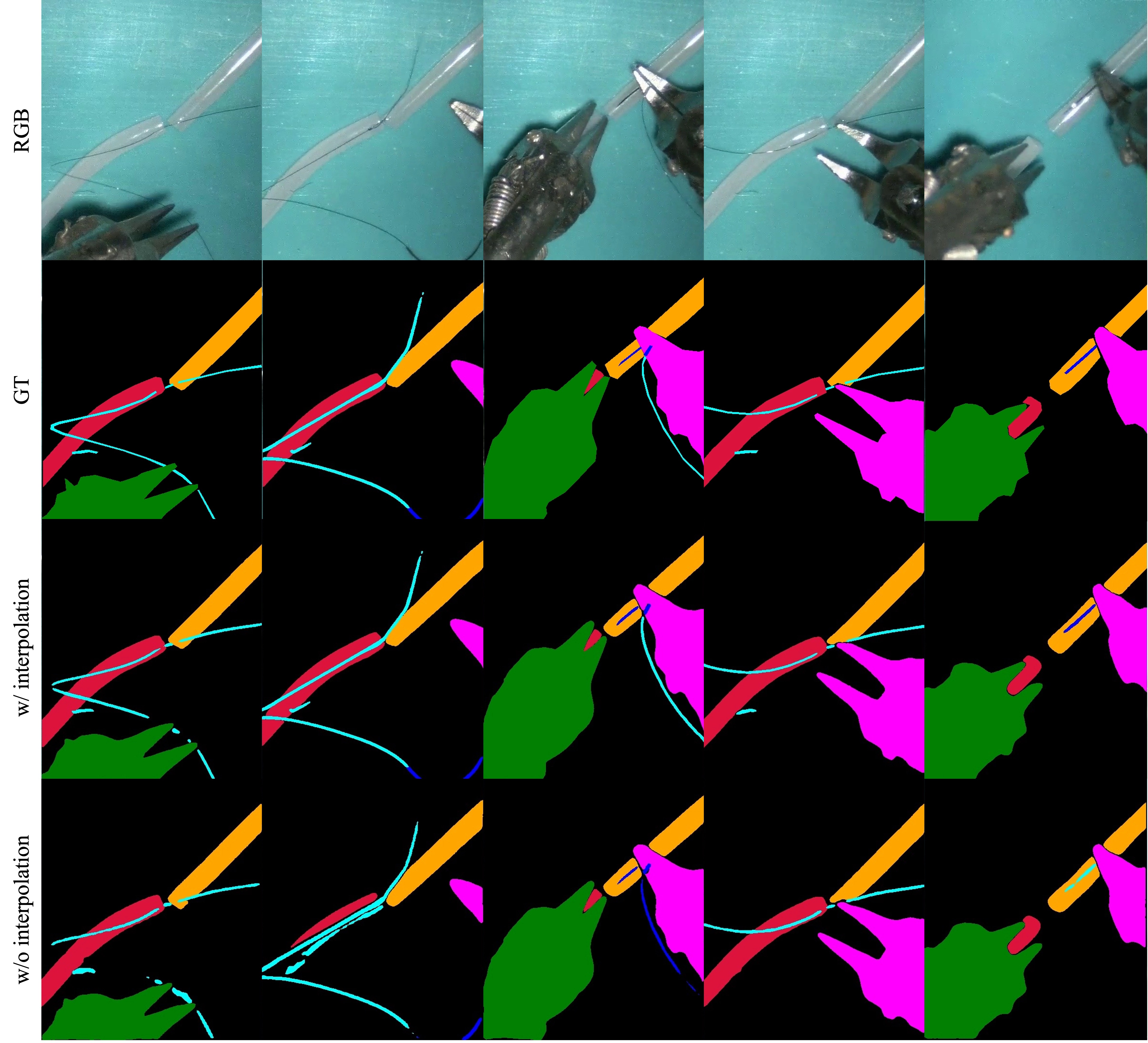}
    \captionof{figure}{Visualization of Instrument Segmentation Results for Comparison on the MISAW Dataset} 
    \label{FIG:misaw_seg}
\end{figure*}
\begin{figure*}
    \centering
    \includegraphics[width=0.8\linewidth]{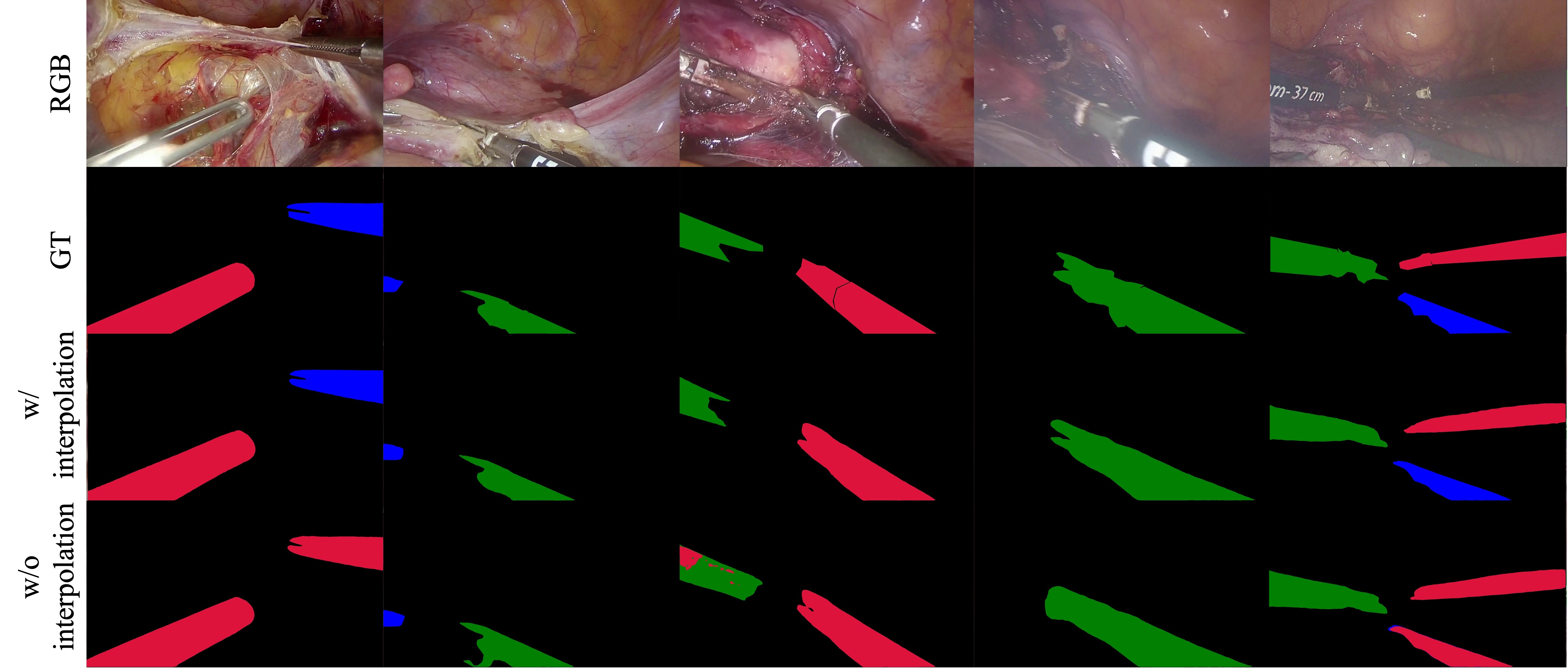}
    \captionof{figure}{Visualization of Instrument Segmentation Results for Comparison on the AutoLaparo Dataset} 
    \label{FIG:autolaparo_seg}
\end{figure*}

\paragraph{MISAW.}
Table~\ref{tab:misaw} reports results on MISAW. Consistent with GraSP, multi-task learning without interpolation degrades spatial task performance, whereas our proposed method achieves the best results across nearly all metrics with particularly strong improvements in workflow recognition and step anticipation. Instrument segmentation also recovers and surpasses the single-task baseline across all metrics, further validating that flow-guided label interpolation effectively resolves annotation imbalance and enables balanced cross-task optimization. Qualitative results in Fig.~\ref{FIG:misaw_seg} further corroborate these findings, showing precise delineation of thin instruments such as suture threads and needles, and robust segmentation under specular reflections and illumination-induced occlusion.

\paragraph{AutoLaparo.}
Table~\ref{tab:autolaparo} reports instrument segmentation results on AutoLaparo. As shown in Table~\ref{tab:autolaparo} and Fig.~\ref{FIG:autolaparo_seg}, the model trained with interpolated pseudo-labels consistently improves over the baseline without interpolation, demonstrating that flow-guided label densification effectively strengthens spatial supervision and improves segmentation performance even in the absence of joint multi-task training. The qualitative results particularly highlight robustness under smoke-induced occlusion, which is a characteristic challenge in laparoscopic surgical environments.

\section{Discussion}

\begin{figure*}
    \centering
    \vspace{-0.15cm}
    \includegraphics[width=0.8\linewidth]{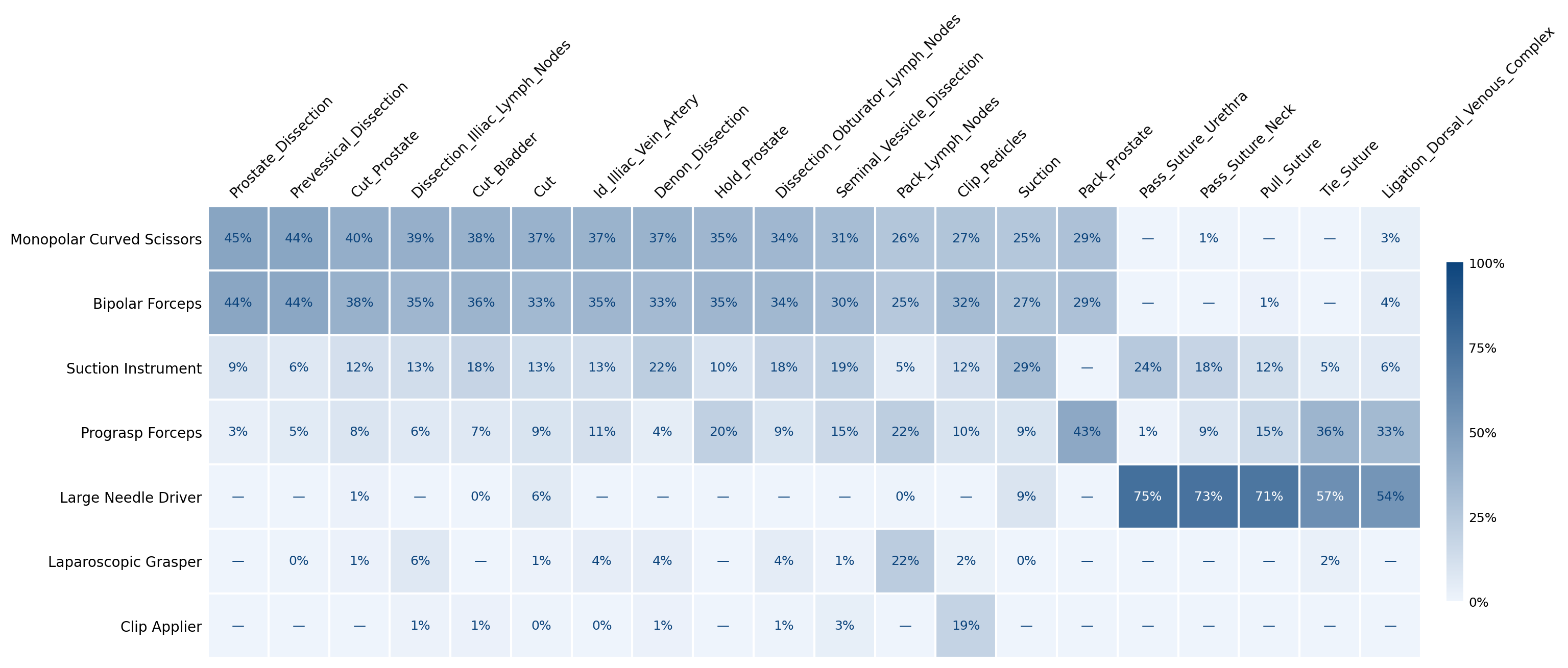}
    \captionof{figure}{Co-occurrence matrix between instrument and step in the GraSP. Each cell indicates the proportion of frames in which a given instrument appears during the corresponding step, normalized per instrument row. The Large Needle Driver exhibits strong co-occurrence with multiple suture-related step categories, while the Clip Applier is predominantly associated with the Clip Pedicles step.}
    \label{FIG:grasp_dis1}
\end{figure*}

\begin{figure}
    \centering
    \vspace{-0.5cm}
    \includegraphics[width=\linewidth]{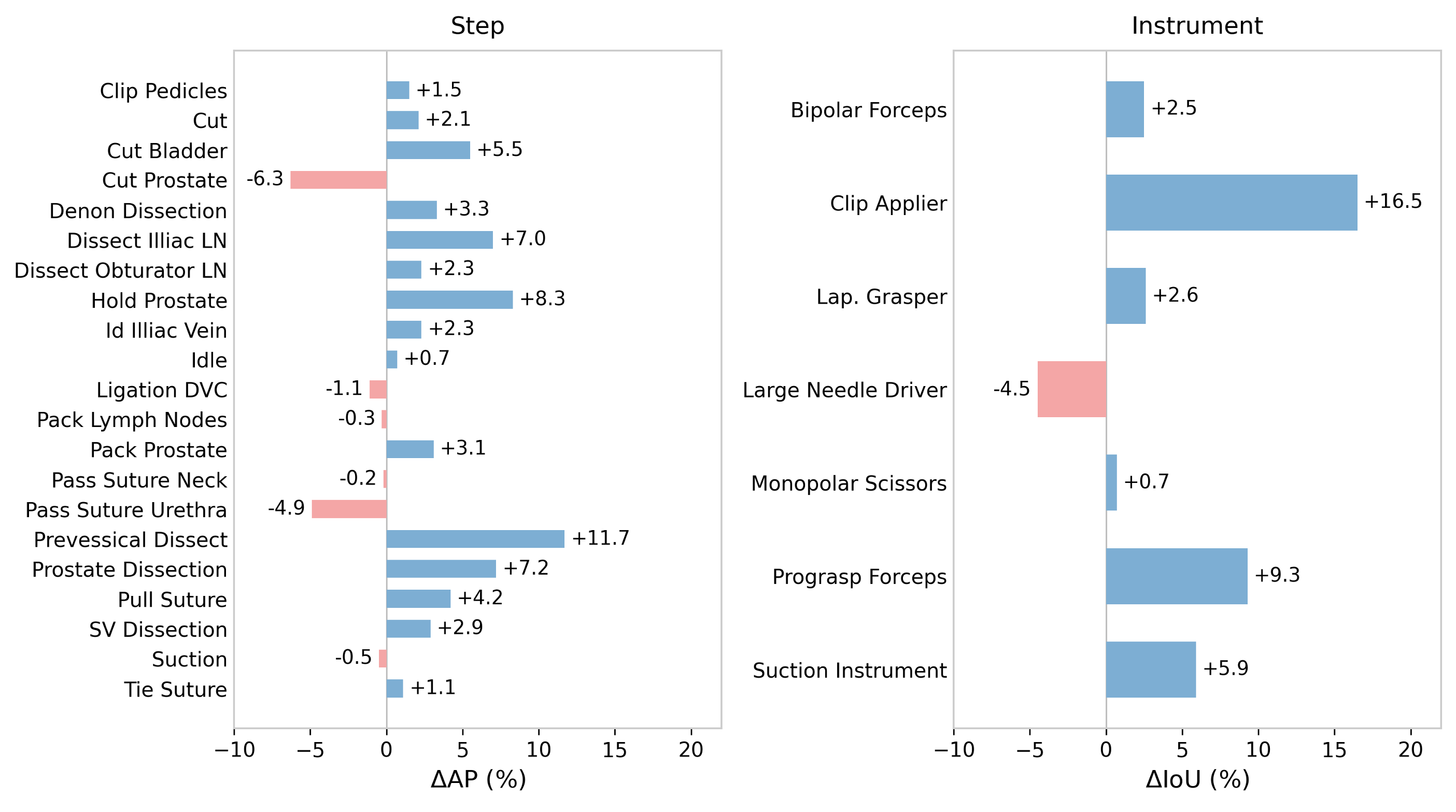}
    \captionof{figure}{Per-class Step recognition AP change($\Delta$AP = w/ inter $-$ w/o inter)(left) and instrument segmentation IoU change ($\Delta$IoU = w/ inter $-$ w/o inter)(right) on GraSP.}
    \label{FIG:grasp_st_inst_ap}
\end{figure}

\paragraph{GraSP.}
The quantitative improvements observed on GraSP provide evidence that instrument segmentation quality directly conditions workflow recognition performance.
In the instrument segmentation task, flow-guided label interpolation yields consistent IoU gains across all instrument categories, as shown in Fig.~\ref{FIG:grasp_st_inst_ap}, with the exception of the Large Needle Driver, which exhibits a slight performance decline.
Analysis of co-occurrence patterns between instrument usage and step categories (Fig.~\ref{FIG:grasp_dis1}) reveals that the Large Needle Driver is strongly correlated with five step categories: \textit{Pass Suture Neck}, \textit{Pass Suture Urethra}, \textit{Cut Prostate}, \textit{Tie Suture}, and \textit{Pull Suture}.
Consistent with this, step recognition performance declines in 6 out of 21 categories under interpolation (Fig.~\ref{FIG:grasp_st_inst_ap}), and notably, 3 of these 6 degraded categories share strong co-occurrence with the Large Needle Driver.
This suggests that the IoU regression of a single instrument can propagate selectively to semantically linked step categories, underscoring the instrument--step dependency embedded in our multi-task framework.

\begin{figure}
    \centering
    \vspace{-0.3cm}
    \includegraphics[width=\linewidth]{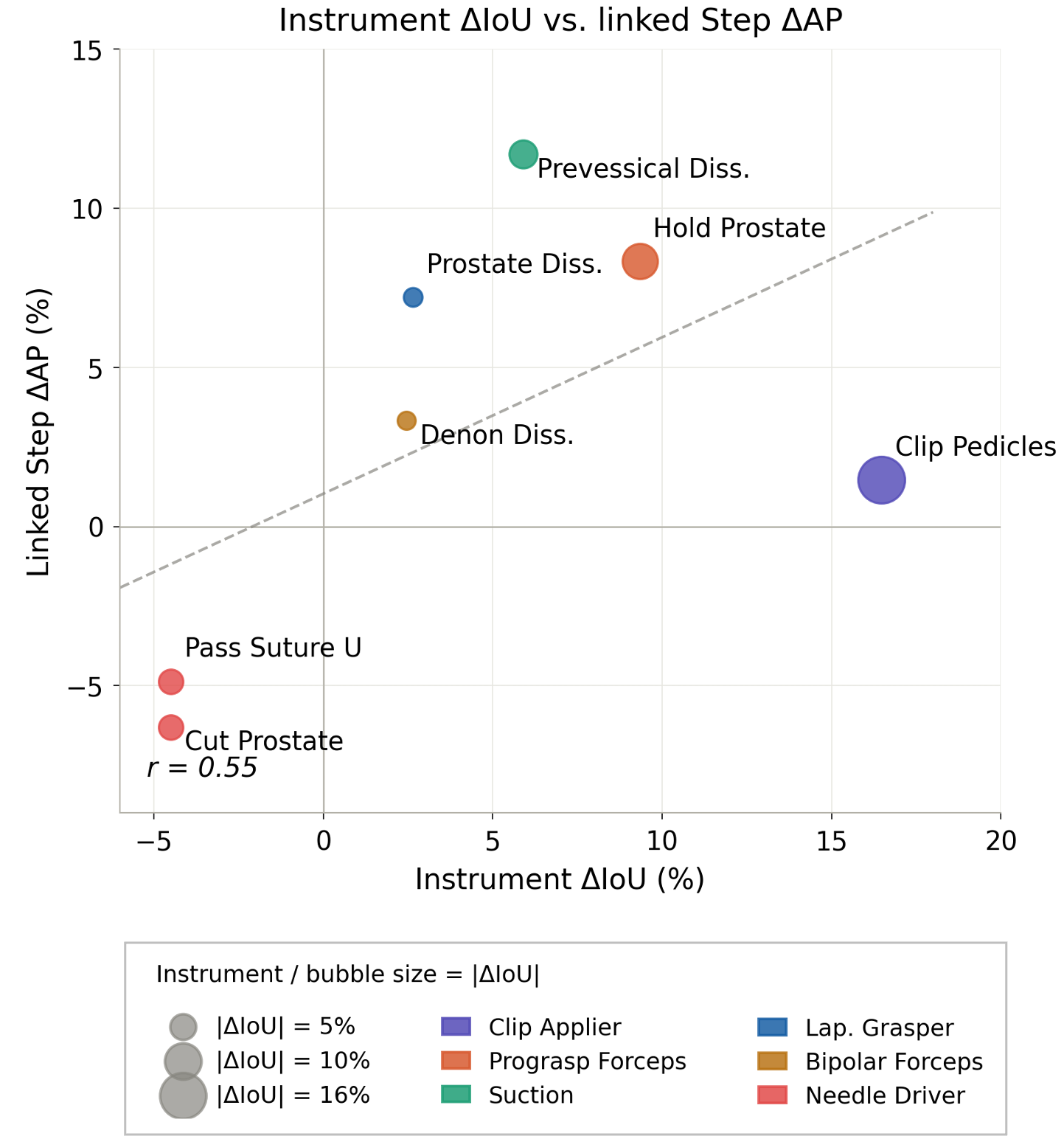}
    \captionof{figure}{Scatter plot of instrument $\Delta$IoU versus linked step $\Delta$AP on GraSP. Each bubble represents an instrument--step pair with strong co-occurrence, and bubble size is proportional to the magnitude of instrument IoU change.}
    \label{FIG:grasp_dis3}
\end{figure}

Conversely, the Clip Applier --- a low-frequency instrument with a baseline IoU of only 42.3\% (w/o interpolation) --- achieves the largest segmentation gain under interpolation ($+16.5$\%p), reflecting that flow-guided densification particularly benefits rare instruments whose sparse annotations are otherwise insufficient for robust learning.
The co-occurrence analysis (Fig.~\ref{FIG:grasp_dis1}) further confirms that the Clip Applier is predominantly associated with the \textit{Clip Pedicles} step, which correspondingly shows a positive AP gain under interpolation.
Taken together, as visualized in Fig.~\ref{FIG:grasp_dis3}, the direction of performance change in step recognition can be anticipated from instrument-level co-occurrence structure, indicating that enriching instrument-level supervision through interpolation propagates beneficial signals to semantically coupled workflow categories.

\begin{figure}
    \centering
    \includegraphics[width=0.8\linewidth]{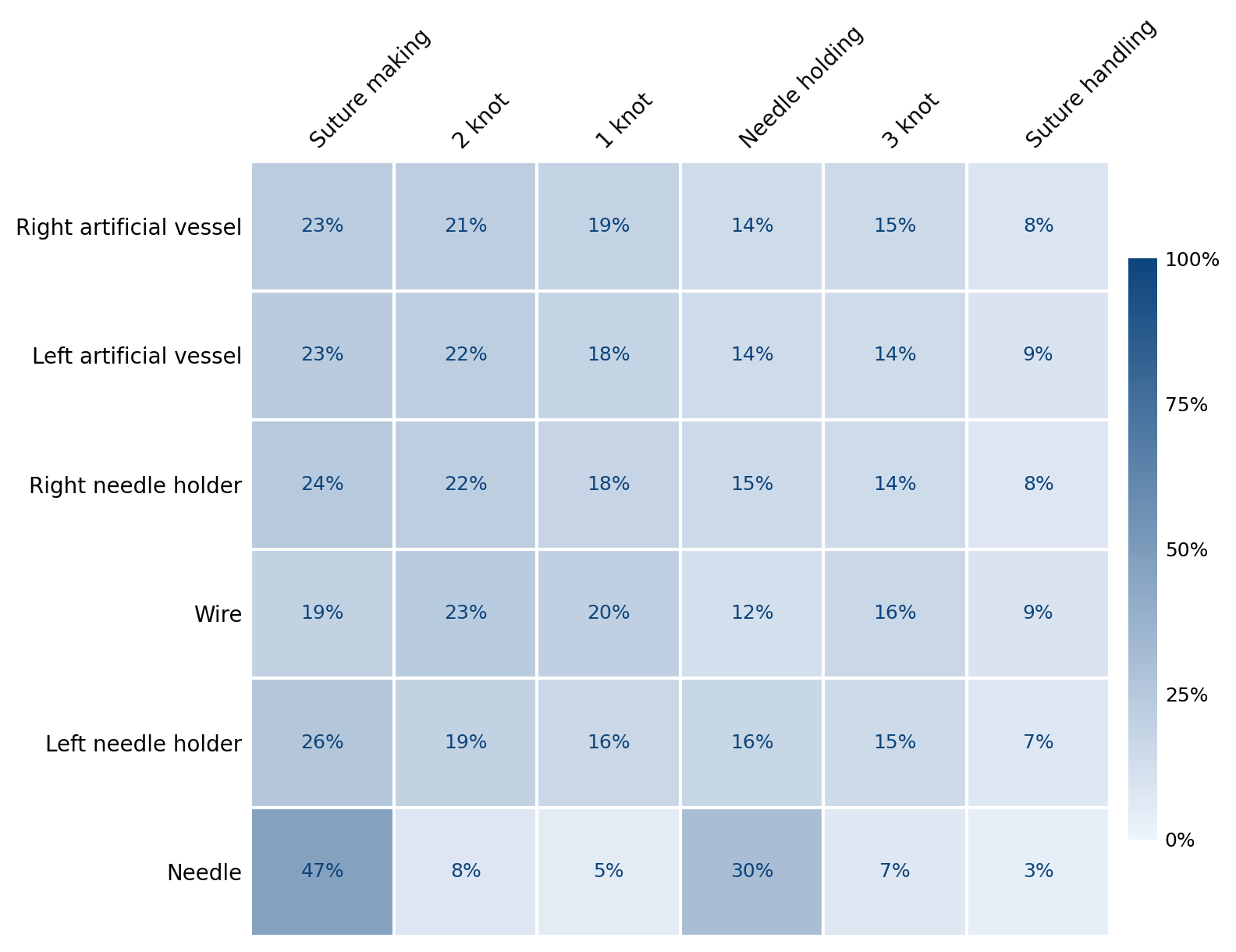}
    \captionof{figure}{Co-occurrence matrix between instrument and step categories in the MISAW. The Needle instrument shows strong association with the Suture Making and Needle Holding categories, reflecting its functional role in suturing-phase procedures.}
    \label{FIG:misaw_dis1}
\end{figure}

\begin{figure}
    \centering
    \vspace{-0.2cm}
    \includegraphics[width=\linewidth]{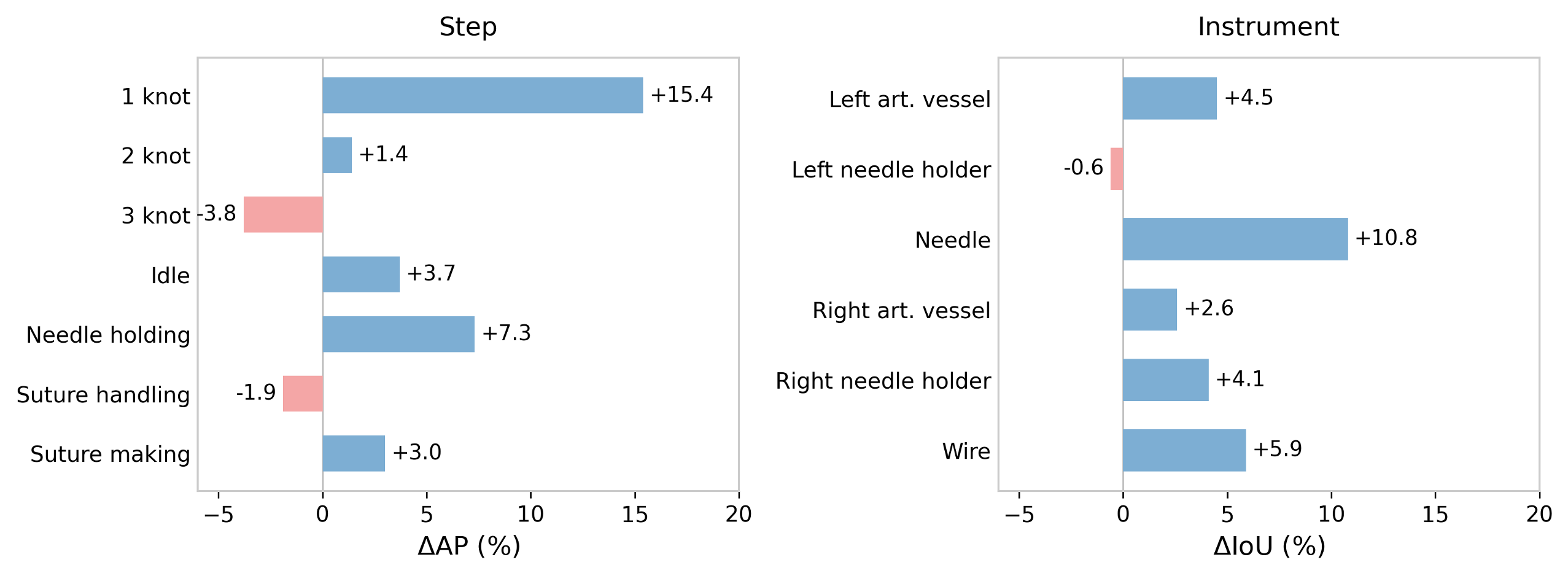}
    \captionof{figure}{Per-class Step recognition AP change($\Delta$AP = w/ inter $-$ w/o inter)(left) and instrument segmentation IoU change ($\Delta$IoU = w/ inter $-$ w/o inter)(right) on MISAW.}
    \label{FIG:misaw_st_inst_ap}
\end{figure}

\begin{figure}
    \centering
    \includegraphics[width=0.9\linewidth]{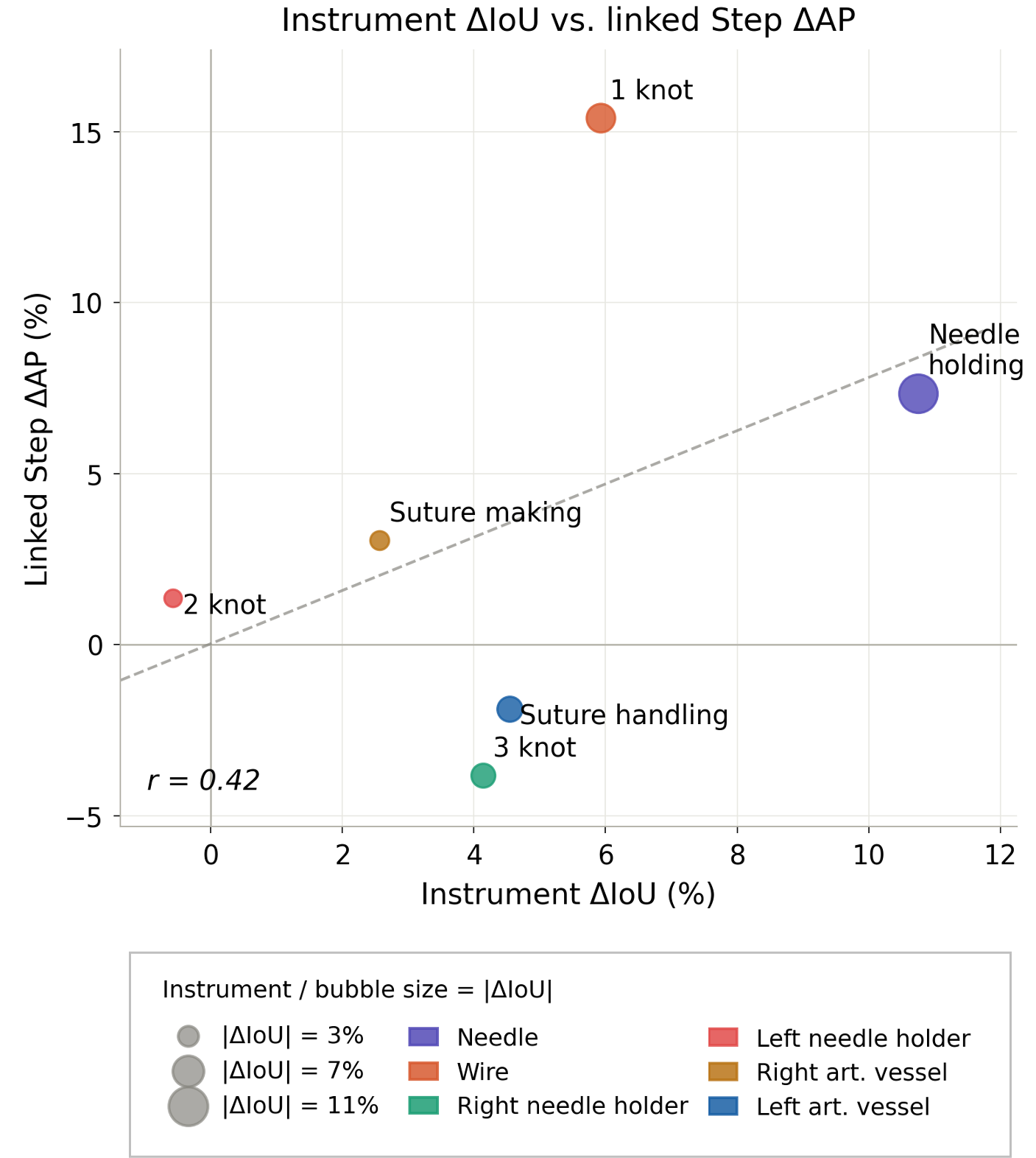}
    \captionof{figure}{Scatter plot of instrument $\Delta$IoU versus linked step $\Delta$AP on MISAW. Bubble size reflects the magnitude of instrument IoU change. The positive correlation confirms that instruments benefiting most from interpolation propagate performance gains to their semantically associated step categories, consistent with findings on GraSP.}
    \label{FIG:misaw_dis3}
\end{figure}

\paragraph{MISAW.}
On MISAW dataset, a similar instrument--step dependency is observed, reinforcing the generalizability of the above findings.
Instrument segmentation improves across all categories under interpolation (Fig.~\ref{FIG:misaw_st_inst_ap}), with the sole exception of the Left Needle Holder, while the Needle class shows the most pronounced gain.
Co-occurrence analysis (Fig.~\ref{FIG:misaw_dis1}) reveals that the Needle instrument is strongly associated with the \textit{Suture Making} and \textit{Needle Holding} step categories, both of which exhibit step AP improvements under interpolation (Fig.~\ref{FIG:misaw_st_inst_ap}), with gains of $+3.1$\%p and $+7.4$\%p, respectively.
The alignment between instrument-level segmentation gains and step-level recognition improvements, further illustrated in Fig.~\ref{FIG:misaw_dis3}, supports the central claim of this work: that instrument spatial information serves as a critical intermediate representation linking low-level perception to high-level workflow understanding, and that flow-guided label interpolation effectively amplifies this linkage by providing denser and more temporally consistent instrument supervision.

\paragraph{Inference Efficiency.}

Beyond accuracy, real-time capability is a prerequisite for intraoperative deployment, where workflow recognition and instrument understanding must keep pace with the surgical video stream.
On the workflow recognition tasks, our framework attains a mean per-clip latency of $22.72 \pm 5.49$~ms, corresponding to a throughput of $44.02$~clips/s at $32$-bit precision on a single GPU, comfortably exceeding the typical $25$--$30$~FPS rate of surgical endoscopic video.
This efficiency is a direct consequence of the two-stage design: the Mask2Former region-proposal network is trained in Stage~1 and subsequently frozen, so that only its precomputed region embeddings are consumed during all-task finetuning and inference.
By decoupling the computationally heavy mask generation from the multi-task heads, the framework avoids redundant per-frame segmentation cost at inference time, enabling joint workflow and region-level prediction at real-time speed without sacrificing the dense spatial grounding provided by the interpolated pseudo-labels.

\section{Conclusion}

Dense temporal supervision for workflow tasks and sparse pixel-level supervision for spatial tasks create an imbalance that undermines joint optimization. To address this, we propose FAROS, a flow-guided label interpolation framework that combines promptable segmentation-based mask propagation 
with optical flow estimation to generate temporally consistent dense pseudo labels from sparse keyframe annotations under challenging surgical conditions. The densified spatial supervision is integrated into a unified Transformer-based multi-task framework with task-dedicated CLS tokens and cross-task conditioning, jointly optimizing surgical phase recognition, step recognition, anticipation, instrument segmentation, and action recognition.
Experiments on GraSP and MISAW benchmarks demonstrate that flow-guided label interpolation is essential for realizing the benefits of multi-task learning: without interpolation, joint training degrades performance relative to single-task baselines, whereas our full model achieves consistent improvements across all tasks. Statistical analysis further reveals that instrument configurations serve as structured priors for workflow recognition, explaining why dense spatial supervision yields particularly large gains in step recognition. We hope this work provides a useful foundation for annotation-efficient holistic surgical scene understanding.

\appendix
\section{Appendix}

\subsection{Effect of Keyframe Interval on Label Interpolation Quality}

Since FAROS generates pseudo-annotations by propagating and correcting masks between sparse annotated keyframes, the keyframe interval is a critical factor in determining interpolation quality. To assess robustness across different annotation densities, we evaluate FAROS on DAVIS 2017 under sparse-GT protocols with keyframe intervals of 20, 30, and 40 frames.
As shown in Table~\ref{tab:davis_stride}, FAROS consistently outperforms the SAM2 baseline across all tested intervals, demonstrating that flow-guided anomaly detection and geometry-driven reprompting provide reliable improvement regardless of keyframe spacing. These results are particularly relevant to our surgical datasets: MISAW and AutoLaparo have keyframe intervals of 30 and 25 frames, respectively, which fall within the range evaluated here and confirm the practical effectiveness of FAROS under typical surgical annotation conditions. While GraSP has a substantially larger interval of approximately 1,000 frames --- a more extreme sparsity condition --- the consistent gains observed across all intervals suggest that the core mechanism of FAROS remains effective even as inter-keyframe distance increases.

\begin{table}[t]
\centering
\caption{Effect of keyframe interval on FAROS label interpolation quality on DAVIS 2017 validation set (sparse-GT protocol). SAM2 baseline and FAROS are compared across different keyframe intervals. $\mathcal{J}$: region similarity; $\mathcal{F}$: contour accuracy.}
\label{tab:davis_stride}
\footnotesize
\renewcommand{\arraystretch}{1.05}
\begin{tabular}{llccc}
\toprule
Interval & Method & $\mathcal{J\&F}$ & $\mathcal{J}$ & $\mathcal{F}$ \\
\midrule
\multirow{2}{*}{20} & SAM2 baseline & 91.5 & 89.1 & 93.9 \\
                    & FAROS (Ours)  & \textbf{93.6} & \textbf{91.2} & \textbf{96.0} \\
\midrule
\multirow{2}{*}{30} & SAM2 baseline & 91.7 & 89.4 & 93.9 \\
                    & FAROS (Ours)  & \textbf{93.3} & \textbf{91.0} & \textbf{95.6} \\
\midrule
\multirow{2}{*}{40} & SAM2 baseline & 91.1 & 88.9 & 93.4 \\
                    & FAROS (Ours)  & \textbf{93.4} & \textbf{91.2} & \textbf{95.7} \\
\bottomrule
\end{tabular}
\end{table}

\subsection{Case Study: FAROS on Challenging Exit/Entry Sequences}

The DAVIS 2017 \textit{drone} subset presents a particularly challenging scenario for mask propagation, characterized by frequent instrument exit and re-entry, rapid object motion, and complex background dynamics --- conditions that closely mirror the challenging surgical environments targeted by FAROS. Since the \textit{drone} subset is not included in the standard DAVIS 2017 validation split, we conduct a separate case study to evaluate FAROS robustness under these conditions.

\begin{figure}
    \centering
    \includegraphics[width=0.9\linewidth]{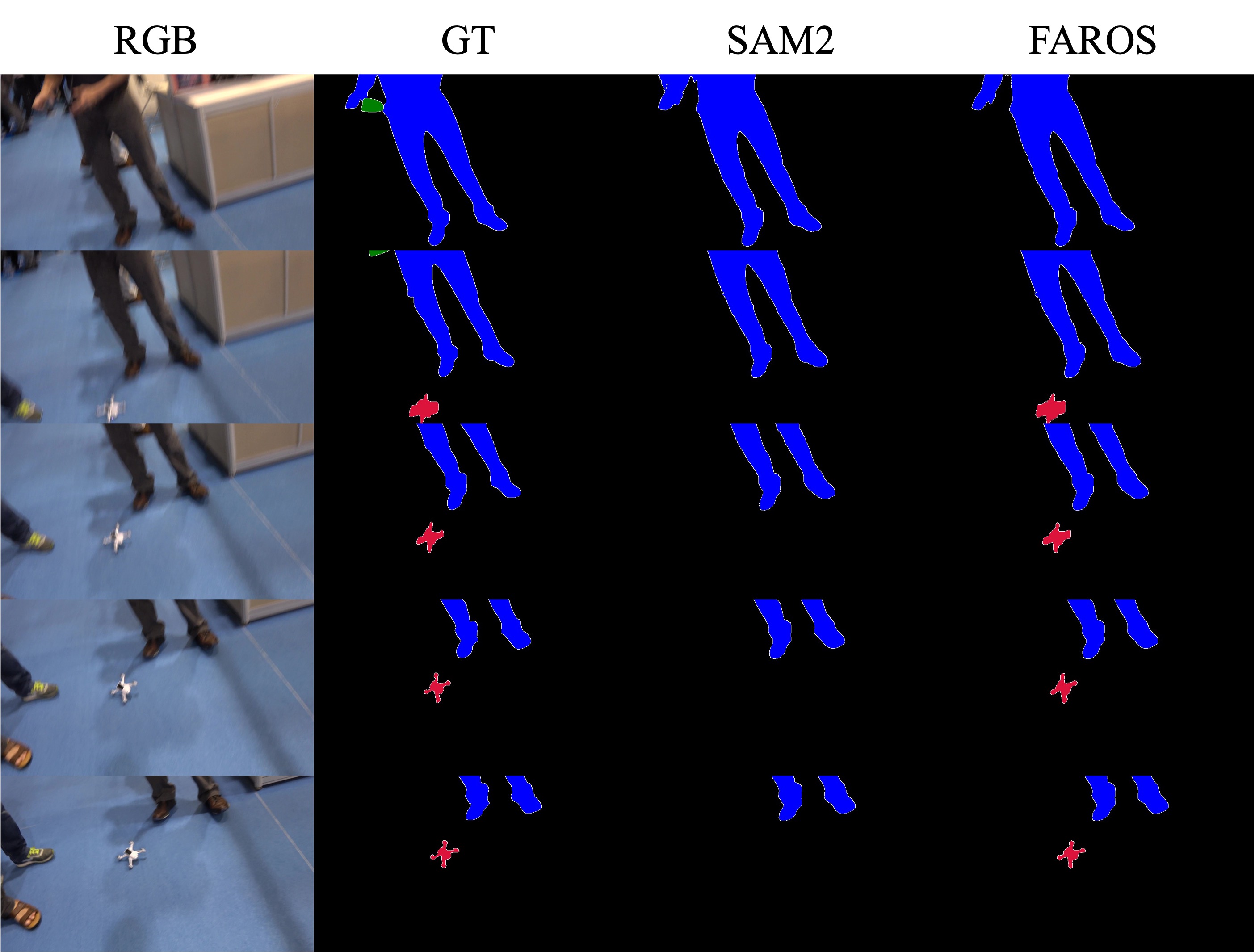}
    \captionof{figure}{Qualitative comparison of mask propagation on the DAVIS 2017 \textit{drone} subset 
    (sparse-GT, interval = 30). Blue and red masks denote the person and drone objects, respectively. As the drone enters the scene, SAM2 baseline fails to detect and track it, producing missing masks throughout the re-entry sequence. FAROS successfully recovers accurate segmentation of both objects via flow-guided reprompting, maintaining temporally consistent masks despite rapid motion and abrupt object appearance.}
    \label{FIG:drone}
\end{figure}

As shown in Table~\ref{tab:davis_drone}, FAROS achieves $\mathcal{J\&F}$ of 0.9308, substantially outperforming the SAM2 baseline (0.8190) by $+11.18\%$p. The particularly large margin in $\mathcal{F}$ ($+11.74\%$p) indicates that flow-guided reprompting is especially effective at recovering accurate object boundaries following exit and re-entry events. These results demonstrate that FAROS maintains robust propagation quality under the most challenging motion conditions, further validating its suitability for surgical video scenarios where instrument exit and occlusion are frequent, as illustrated in Fig.~\ref{FIG:drone}.

\begin{table}[h]
\centering
\caption{Case study on the DAVIS 2017 \textit{drone} subset (sparse-GT, interval = 30). The \textit{drone} subset is excluded from the standard validation split and evaluated separately as a challenging case study involving frequent object exit and re-entry.}
\label{tab:davis_drone}
\footnotesize
\renewcommand{\arraystretch}{1.05}
\begin{tabular}{lccc}
\toprule
Method & $\mathcal{J\&F}$ & $\mathcal{J}$ & $\mathcal{F}$ \\
\midrule
SAM2 baseline & 81.9 & 80.3 & 83.5 \\
FAROS (Ours)  & \textbf{93.0} & \textbf{90.9} & \textbf{95.2} \\
\bottomrule
\end{tabular}
\end{table}

\printcredits

\bibliographystyle{cas-model2-names}

\bibliography{main} 

\begin{thebibliography}{74}
\expandafter\ifx\csname natexlab\endcsname\relax\def\natexlab#1{#1}\fi
\providecommand{\url}[1]{\texttt{#1}}
\providecommand{\href}[2]{#2}
\providecommand{\path}[1]{#1}
\providecommand{\DOIprefix}{doi:}
\providecommand{\ArXivprefix}{arXiv:}
\providecommand{\URLprefix}{URL: }
\providecommand{\Pubmedprefix}{pmid:}
\providecommand{\doi}[1]{\href{http://dx.doi.org/#1}{\path{#1}}}
\providecommand{\Pubmed}[1]{\href{pmid:#1}{\path{#1}}}
\providecommand{\bibinfo}[2]{#2}
\ifx\xfnm\relax \def\xfnm[#1]{\unskip,\space#1}\fi
\bibitem[{Ahmed et~al.(2024)Ahmed, Yousef, Ahmed, Ali, Mahboob, Ali, Shah, Aboumarzouk, Al~Ansari and Balakrishnan}]{ahmed2024deep}
\bibinfo{author}{Ahmed, F.A.}, \bibinfo{author}{Yousef, M.}, \bibinfo{author}{Ahmed, M.A.}, \bibinfo{author}{Ali, H.O.}, \bibinfo{author}{Mahboob, A.}, \bibinfo{author}{Ali, H.}, \bibinfo{author}{Shah, Z.}, \bibinfo{author}{Aboumarzouk, O.}, \bibinfo{author}{Al~Ansari, A.}, \bibinfo{author}{Balakrishnan, S.}, \bibinfo{year}{2024}.
\newblock \bibinfo{title}{Deep learning for surgical instrument recognition and segmentation in robotic-assisted surgeries: a systematic review}.
\newblock \bibinfo{journal}{Artificial Intelligence Review} \bibinfo{volume}{58}, \bibinfo{pages}{1}.
\bibitem[{Alabi et~al.(2025)Alabi, Vercauteren and Shi}]{alabi2025multitask}
\bibinfo{author}{Alabi, O.}, \bibinfo{author}{Vercauteren, T.}, \bibinfo{author}{Shi, M.}, \bibinfo{year}{2025}.
\newblock \bibinfo{title}{Multitask learning in minimally invasive surgical vision: A review}.
\newblock \bibinfo{journal}{Medical Image Analysis} \bibinfo{volume}{101}, \bibinfo{pages}{103480}.
\bibitem[{Allan et~al.(2020)Allan, Kondo, Bodenstedt, Leger, Kadkhodamohammadi, Luengo, Fuentes, Flouty, Mohammed, Pedersen et~al.}]{allan20202018}
\bibinfo{author}{Allan, M.}, \bibinfo{author}{Kondo, S.}, \bibinfo{author}{Bodenstedt, S.}, \bibinfo{author}{Leger, S.}, \bibinfo{author}{Kadkhodamohammadi, R.}, \bibinfo{author}{Luengo, I.}, \bibinfo{author}{Fuentes, F.}, \bibinfo{author}{Flouty, E.}, \bibinfo{author}{Mohammed, A.}, \bibinfo{author}{Pedersen, M.}, et~al., \bibinfo{year}{2020}.
\newblock \bibinfo{title}{2018 robotic scene segmentation challenge}.
\newblock \bibinfo{journal}{arXiv preprint arXiv:2001.11190} .
\bibitem[{Antonelli et~al.(2022)Antonelli, Reinke, Bakas, Farahani, Kopp-Schneider, Landman, Litjens, Menze, Ronneberger, Summers et~al.}]{antonelli2022medical}
\bibinfo{author}{Antonelli, M.}, \bibinfo{author}{Reinke, A.}, \bibinfo{author}{Bakas, S.}, \bibinfo{author}{Farahani, K.}, \bibinfo{author}{Kopp-Schneider, A.}, \bibinfo{author}{Landman, B.A.}, \bibinfo{author}{Litjens, G.}, \bibinfo{author}{Menze, B.}, \bibinfo{author}{Ronneberger, O.}, \bibinfo{author}{Summers, R.M.}, et~al., \bibinfo{year}{2022}.
\newblock \bibinfo{title}{The medical segmentation decathlon}.
\newblock \bibinfo{journal}{Nature communications} \bibinfo{volume}{13}, \bibinfo{pages}{4128}.
\bibitem[{Ayobi et~al.(2024)Ayobi, Rodr{\'\i}guez, P{\'e}rez, Hern{\'a}ndez, Aparicio, Dessevres, Pe{\~n}a, Santander, Caicedo, Fern{\'a}ndez et~al.}]{ayobi2024pixel}
\bibinfo{author}{Ayobi, N.}, \bibinfo{author}{Rodr{\'\i}guez, S.}, \bibinfo{author}{P{\'e}rez, A.}, \bibinfo{author}{Hern{\'a}ndez, I.}, \bibinfo{author}{Aparicio, N.}, \bibinfo{author}{Dessevres, E.}, \bibinfo{author}{Pe{\~n}a, S.}, \bibinfo{author}{Santander, J.}, \bibinfo{author}{Caicedo, J.I.}, \bibinfo{author}{Fern{\'a}ndez, N.}, et~al., \bibinfo{year}{2024}.
\newblock \bibinfo{title}{Pixel-wise recognition for holistic surgical scene understanding}.
\newblock \bibinfo{journal}{arXiv preprint arXiv:2401.11174} .
\bibitem[{Baghbaderani et~al.(2024)Baghbaderani, Li, Wang and Qi}]{baghbaderani2024temporally}
\bibinfo{author}{Baghbaderani, R.K.}, \bibinfo{author}{Li, Y.}, \bibinfo{author}{Wang, S.}, \bibinfo{author}{Qi, H.}, \bibinfo{year}{2024}.
\newblock \bibinfo{title}{Temporally-consistent video semantic segmentation with bidirectional occlusion-guided feature propagation}, in: \bibinfo{booktitle}{Proceedings of the IEEE/CVF Winter Conference on Applications of Computer Vision}, pp. \bibinfo{pages}{685--695}.
\bibitem[{Bai et~al.(2017)Bai, Oktay, Sinclair, Suzuki, Rajchl, Tarroni, Glocker, King, Matthews and Rueckert}]{bai2017semi}
\bibinfo{author}{Bai, W.}, \bibinfo{author}{Oktay, O.}, \bibinfo{author}{Sinclair, M.}, \bibinfo{author}{Suzuki, H.}, \bibinfo{author}{Rajchl, M.}, \bibinfo{author}{Tarroni, G.}, \bibinfo{author}{Glocker, B.}, \bibinfo{author}{King, A.}, \bibinfo{author}{Matthews, P.M.}, \bibinfo{author}{Rueckert, D.}, \bibinfo{year}{2017}.
\newblock \bibinfo{title}{Semi-supervised learning for network-based cardiac mr image segmentation}, in: \bibinfo{booktitle}{International Conference on Medical Image Computing and Computer-Assisted Intervention}, \bibinfo{organization}{Springer}. pp. \bibinfo{pages}{253--260}.
\bibitem[{Caruana(1997)}]{caruana1997multitask}
\bibinfo{author}{Caruana, R.}, \bibinfo{year}{1997}.
\newblock \bibinfo{title}{Multitask learning}.
\newblock \bibinfo{journal}{Machine learning} \bibinfo{volume}{28}, \bibinfo{pages}{41--75}.
\bibitem[{Chen and Poullis(2022)}]{chen2022motion}
\bibinfo{author}{Chen, Q.}, \bibinfo{author}{Poullis, C.}, \bibinfo{year}{2022}.
\newblock \bibinfo{title}{Motion estimation for large displacements and deformations}.
\newblock \bibinfo{journal}{Scientific Reports} \bibinfo{volume}{12}, \bibinfo{pages}{19721}.
\bibitem[{Cheng et~al.(2022)Cheng, Misra, Schwing, Kirillov and Girdhar}]{mask2former}
\bibinfo{author}{Cheng, B.}, \bibinfo{author}{Misra, I.}, \bibinfo{author}{Schwing, A.G.}, \bibinfo{author}{Kirillov, A.}, \bibinfo{author}{Girdhar, R.}, \bibinfo{year}{2022}.
\newblock \bibinfo{title}{Masked-attention mask transformer for universal image segmentation}, in: \bibinfo{booktitle}{Proceedings of the IEEE/CVF conference on computer vision and pattern recognition}, pp. \bibinfo{pages}{1290--1299}.
\bibitem[{Cheng and Schwing(2022)}]{Xmem}
\bibinfo{author}{Cheng, H.K.}, \bibinfo{author}{Schwing, A.G.}, \bibinfo{year}{2022}.
\newblock \bibinfo{title}{Xmem: Long-term video object segmentation with an atkinson-shiffrin memory model}, in: \bibinfo{booktitle}{European conference on computer vision}, \bibinfo{organization}{Springer}. pp. \bibinfo{pages}{640--658}.
\bibitem[{Cheng et~al.(2021)Cheng, Tai and Tang}]{STCN}
\bibinfo{author}{Cheng, H.K.}, \bibinfo{author}{Tai, Y.W.}, \bibinfo{author}{Tang, C.K.}, \bibinfo{year}{2021}.
\newblock \bibinfo{title}{Rethinking space-time networks with improved memory coverage for efficient video object segmentation}.
\newblock \bibinfo{journal}{Advances in neural information processing systems} \bibinfo{volume}{34}, \bibinfo{pages}{11781--11794}.
\bibitem[{Chuchulo and Ali(2023)}]{chuchulo2023robotic}
\bibinfo{author}{Chuchulo, A.}, \bibinfo{author}{Ali, A.}, \bibinfo{year}{2023}.
\newblock \bibinfo{title}{Is robotic-assisted surgery better?}
\newblock \bibinfo{journal}{AMA Journal of Ethics} \bibinfo{volume}{25}, \bibinfo{pages}{598--604}.
\bibitem[{Crawshaw(2020)}]{crawshaw2020multi}
\bibinfo{author}{Crawshaw, M.}, \bibinfo{year}{2020}.
\newblock \bibinfo{title}{Multi-task learning with deep neural networks: A survey}.
\newblock \bibinfo{journal}{arXiv preprint arXiv:2009.09796} .
\bibitem[{Czempiel et~al.(2020)Czempiel, Paschali, Keicher, Simson, Feussner, Kim and Navab}]{czempiel2020tecno}
\bibinfo{author}{Czempiel, T.}, \bibinfo{author}{Paschali, M.}, \bibinfo{author}{Keicher, M.}, \bibinfo{author}{Simson, W.}, \bibinfo{author}{Feussner, H.}, \bibinfo{author}{Kim, S.T.}, \bibinfo{author}{Navab, N.}, \bibinfo{year}{2020}.
\newblock \bibinfo{title}{Tecno: Surgical phase recognition with multi-stage temporal convolutional networks}, in: \bibinfo{booktitle}{International conference on medical image computing and computer-assisted intervention}, \bibinfo{organization}{Springer}. pp. \bibinfo{pages}{343--352}.
\bibitem[{Demir et~al.(2023)Demir, Schieber, Weise, Roth, May, Maier and Yang}]{demir2023deep}
\bibinfo{author}{Demir, K.C.}, \bibinfo{author}{Schieber, H.}, \bibinfo{author}{Weise, T.}, \bibinfo{author}{Roth, D.}, \bibinfo{author}{May, M.}, \bibinfo{author}{Maier, A.}, \bibinfo{author}{Yang, S.H.}, \bibinfo{year}{2023}.
\newblock \bibinfo{title}{Deep learning in surgical workflow analysis: a review of phase and step recognition}.
\newblock \bibinfo{journal}{IEEE Journal of Biomedical and Health Informatics} \bibinfo{volume}{27}, \bibinfo{pages}{5405--5417}.
\bibitem[{Everingham et~al.(2015)Everingham, Eslami, Van~Gool, Williams, Winn and Zisserman}]{everingham2015pascal}
\bibinfo{author}{Everingham, M.}, \bibinfo{author}{Eslami, S.A.}, \bibinfo{author}{Van~Gool, L.}, \bibinfo{author}{Williams, C.K.}, \bibinfo{author}{Winn, J.}, \bibinfo{author}{Zisserman, A.}, \bibinfo{year}{2015}.
\newblock \bibinfo{title}{The pascal visual object classes challenge: A retrospective}.
\newblock \bibinfo{journal}{International journal of computer vision} \bibinfo{volume}{111}, \bibinfo{pages}{98--136}.
\bibitem[{Fan et~al.(2021)Fan, Xiong, Mangalam, Li, Yan, Malik and Feichtenhofer}]{mvit}
\bibinfo{author}{Fan, H.}, \bibinfo{author}{Xiong, B.}, \bibinfo{author}{Mangalam, K.}, \bibinfo{author}{Li, Y.}, \bibinfo{author}{Yan, Z.}, \bibinfo{author}{Malik, J.}, \bibinfo{author}{Feichtenhofer, C.}, \bibinfo{year}{2021}.
\newblock \bibinfo{title}{Multiscale vision transformers}, in: \bibinfo{booktitle}{IEEE International Conference on Computer Vision}.
\newblock \DOIprefix\doi{10.1109/ICCV48922.2021.00675}.
\bibitem[{Gao et~al.(2021)Gao, Jin, Long, Dou and Heng}]{gao2021trans}
\bibinfo{author}{Gao, X.}, \bibinfo{author}{Jin, Y.}, \bibinfo{author}{Long, Y.}, \bibinfo{author}{Dou, Q.}, \bibinfo{author}{Heng, P.A.}, \bibinfo{year}{2021}.
\newblock \bibinfo{title}{Trans-svnet: Accurate phase recognition from surgical videos via hybrid embedding aggregation transformer}, in: \bibinfo{booktitle}{International conference on medical image computing and computer-assisted intervention}, \bibinfo{organization}{Springer}. pp. \bibinfo{pages}{593--603}.
\bibitem[{Gu et~al.(2017)Gu, Sun, Vijayanarasimhan, Pantofaru, Ross, Toderici, Li, Ricco, Sukthankar, Schmid and Malik}]{gu2017}
\bibinfo{author}{Gu, C.}, \bibinfo{author}{Sun, C.}, \bibinfo{author}{Vijayanarasimhan, S.}, \bibinfo{author}{Pantofaru, C.}, \bibinfo{author}{Ross, D.A.}, \bibinfo{author}{Toderici, G.}, \bibinfo{author}{Li, Y.}, \bibinfo{author}{Ricco, S.}, \bibinfo{author}{Sukthankar, R.}, \bibinfo{author}{Schmid, C.}, \bibinfo{author}{Malik, J.}, \bibinfo{year}{2017}.
\newblock \bibinfo{title}{Ava: A video dataset of spatio-temporally localized atomic visual actions}, in: \bibinfo{booktitle}{2018 IEEE/CVF Conference on Computer Vision and Pattern Recognition}.
\newblock \DOIprefix\doi{10.1109/CVPR.2018.00633}.
\bibitem[{Handa et~al.(2024)Handa, Gaidhane and Choudhari}]{handa2024role}
\bibinfo{author}{Handa, A.}, \bibinfo{author}{Gaidhane, A.}, \bibinfo{author}{Choudhari, S.G.}, \bibinfo{year}{2024}.
\newblock \bibinfo{title}{Role of robotic-assisted surgery in public health: its advantages and challenges}.
\newblock \bibinfo{journal}{Cureus} \bibinfo{volume}{16}.
\bibitem[{Huaulm{\'e} et~al.(2021)Huaulm{\'e}, Sarikaya, Le~Mut, Despinoy, Long, Dou, Chng, Lin, Kondo, Bravo-S{\'a}nchez et~al.}]{huaulme2021micro}
\bibinfo{author}{Huaulm{\'e}, A.}, \bibinfo{author}{Sarikaya, D.}, \bibinfo{author}{Le~Mut, K.}, \bibinfo{author}{Despinoy, F.}, \bibinfo{author}{Long, Y.}, \bibinfo{author}{Dou, Q.}, \bibinfo{author}{Chng, C.B.}, \bibinfo{author}{Lin, W.}, \bibinfo{author}{Kondo, S.}, \bibinfo{author}{Bravo-S{\'a}nchez, L.}, et~al., \bibinfo{year}{2021}.
\newblock \bibinfo{title}{Micro-surgical anastomose workflow recognition challenge report}.
\newblock \bibinfo{journal}{Computer Methods and Programs in Biomedicine} \bibinfo{volume}{212}, \bibinfo{pages}{106452}.
\bibitem[{Jeong et~al.(2025)Jeong, Kim and Park}]{misaw_segdata}
\bibinfo{author}{Jeong, T.K.}, \bibinfo{author}{Kim, G.}, \bibinfo{author}{Park, J.}, \bibinfo{year}{2025}.
\newblock \bibinfo{title}{Microsurgical instrument segmentation for robot-assisted surgery}.
\newblock \bibinfo{journal}{arXiv preprint arXiv:2509.11727} .
\bibitem[{Jin et~al.(2019)Jin, Cheng, Dou and Heng}]{jin2019incorporating}
\bibinfo{author}{Jin, Y.}, \bibinfo{author}{Cheng, K.}, \bibinfo{author}{Dou, Q.}, \bibinfo{author}{Heng, P.A.}, \bibinfo{year}{2019}.
\newblock \bibinfo{title}{Incorporating temporal prior from motion flow for instrument segmentation in minimally invasive surgery video}, in: \bibinfo{booktitle}{International conference on medical image computing and computer-assisted intervention}, \bibinfo{organization}{Springer}. pp. \bibinfo{pages}{440--448}.
\bibitem[{Jin et~al.(2017)Jin, Dou, Chen, Yu, Qin, Fu and Heng}]{jin2017sv}
\bibinfo{author}{Jin, Y.}, \bibinfo{author}{Dou, Q.}, \bibinfo{author}{Chen, H.}, \bibinfo{author}{Yu, L.}, \bibinfo{author}{Qin, J.}, \bibinfo{author}{Fu, C.W.}, \bibinfo{author}{Heng, P.A.}, \bibinfo{year}{2017}.
\newblock \bibinfo{title}{Sv-rcnet: workflow recognition from surgical videos using recurrent convolutional network}.
\newblock \bibinfo{journal}{IEEE transactions on medical imaging} \bibinfo{volume}{37}, \bibinfo{pages}{1114--1126}.
\bibitem[{Jin et~al.(2020)Jin, Li, Dou, Chen, Qin, Fu and Heng}]{jin2020multi}
\bibinfo{author}{Jin, Y.}, \bibinfo{author}{Li, H.}, \bibinfo{author}{Dou, Q.}, \bibinfo{author}{Chen, H.}, \bibinfo{author}{Qin, J.}, \bibinfo{author}{Fu, C.W.}, \bibinfo{author}{Heng, P.A.}, \bibinfo{year}{2020}.
\newblock \bibinfo{title}{Multi-task recurrent convolutional network with correlation loss for surgical video analysis}.
\newblock \bibinfo{journal}{Medical image analysis} \bibinfo{volume}{59}, \bibinfo{pages}{101572}.
\bibitem[{Kirillov et~al.(2023)Kirillov, Mintun, Ravi, Mao, Rolland, Gustafson, Xiao, Whitehead, Berg, Lo et~al.}]{sam}
\bibinfo{author}{Kirillov, A.}, \bibinfo{author}{Mintun, E.}, \bibinfo{author}{Ravi, N.}, \bibinfo{author}{Mao, H.}, \bibinfo{author}{Rolland, C.}, \bibinfo{author}{Gustafson, L.}, \bibinfo{author}{Xiao, T.}, \bibinfo{author}{Whitehead, S.}, \bibinfo{author}{Berg, A.C.}, \bibinfo{author}{Lo, W.Y.}, et~al., \bibinfo{year}{2023}.
\newblock \bibinfo{title}{Segment anything}, in: \bibinfo{booktitle}{Proceedings of the IEEE/CVF international conference on computer vision}, pp. \bibinfo{pages}{4015--4026}.
\bibitem[{Laina et~al.(2017)Laina, Rieke, Rupprecht, Vizca{\'\i}no, Eslami, Tombari and Navab}]{laina2017concurrent}
\bibinfo{author}{Laina, I.}, \bibinfo{author}{Rieke, N.}, \bibinfo{author}{Rupprecht, C.}, \bibinfo{author}{Vizca{\'\i}no, J.P.}, \bibinfo{author}{Eslami, A.}, \bibinfo{author}{Tombari, F.}, \bibinfo{author}{Navab, N.}, \bibinfo{year}{2017}.
\newblock \bibinfo{title}{Concurrent segmentation and localization for tracking of surgical instruments}, in: \bibinfo{booktitle}{International conference on medical image computing and computer-assisted intervention}, \bibinfo{organization}{Springer}. pp. \bibinfo{pages}{664--672}.
\bibitem[{Lalys and Jannin(2014)}]{lalys2014surgical}
\bibinfo{author}{Lalys, F.}, \bibinfo{author}{Jannin, P.}, \bibinfo{year}{2014}.
\newblock \bibinfo{title}{Surgical process modelling: a review}.
\newblock \bibinfo{journal}{International journal of computer assisted radiology and surgery} \bibinfo{volume}{9}, \bibinfo{pages}{495--511}.
\bibitem[{Lee et~al.(2013)}]{lee2013pseudo}
\bibinfo{author}{Lee, D.H.}, et~al., \bibinfo{year}{2013}.
\newblock \bibinfo{title}{Pseudo-label: The simple and efficient semi-supervised learning method for deep neural networks}, in: \bibinfo{booktitle}{Workshop on challenges in representation learning, ICML}, \bibinfo{organization}{Atlanta}. p. \bibinfo{pages}{896}.
\bibitem[{Li et~al.(2022)Li, Hu, Xiong, Zhang, Pan and Liu}]{RDE}
\bibinfo{author}{Li, M.}, \bibinfo{author}{Hu, L.}, \bibinfo{author}{Xiong, Z.}, \bibinfo{author}{Zhang, B.}, \bibinfo{author}{Pan, P.}, \bibinfo{author}{Liu, D.}, \bibinfo{year}{2022}.
\newblock \bibinfo{title}{Recurrent dynamic embedding for video object segmentation}, in: \bibinfo{booktitle}{Proceedings of the IEEE/CVF Conference on Computer Vision and Pattern Recognition}, pp. \bibinfo{pages}{1332--1341}.
\bibitem[{Li et~al.(2016)Li, Cosker and Brown}]{li2016drift}
\bibinfo{author}{Li, W.}, \bibinfo{author}{Cosker, D.}, \bibinfo{author}{Brown, M.}, \bibinfo{year}{2016}.
\newblock \bibinfo{title}{Drift robust non-rigid optical flow enhancement for long sequences}.
\newblock \bibinfo{journal}{Journal of Intelligent \& Fuzzy Systems} \bibinfo{volume}{31}, \bibinfo{pages}{2583--2595}.
\bibitem[{Li et~al.(2024)Li, Zhao, Li and Li}]{li2024deep}
\bibinfo{author}{Li, Y.}, \bibinfo{author}{Zhao, Z.}, \bibinfo{author}{Li, R.}, \bibinfo{author}{Li, F.}, \bibinfo{year}{2024}.
\newblock \bibinfo{title}{Deep learning for surgical workflow analysis: a survey of progresses, limitations, and trends}.
\newblock \bibinfo{journal}{Artificial Intelligence Review} \bibinfo{volume}{57}, \bibinfo{pages}{291}.
\bibitem[{Liu et~al.(2024)Liu, Zhang, Wu, Hong and Jin}]{liu2024surgical}
\bibinfo{author}{Liu, H.}, \bibinfo{author}{Zhang, E.}, \bibinfo{author}{Wu, J.}, \bibinfo{author}{Hong, M.}, \bibinfo{author}{Jin, Y.}, \bibinfo{year}{2024}.
\newblock \bibinfo{title}{Surgical sam 2: Real-time segment anything in surgical video by efficient frame pruning}.
\newblock \bibinfo{journal}{arXiv preprint arXiv:2408.07931} .
\bibitem[{Liu et~al.(2021)Liu, Lin, Cao, Hu, Wei, Zhang, Lin and Guo}]{swin}
\bibinfo{author}{Liu, Z.}, \bibinfo{author}{Lin, Y.}, \bibinfo{author}{Cao, Y.}, \bibinfo{author}{Hu, H.}, \bibinfo{author}{Wei, Y.}, \bibinfo{author}{Zhang, Z.}, \bibinfo{author}{Lin, S.}, \bibinfo{author}{Guo, B.}, \bibinfo{year}{2021}.
\newblock \bibinfo{title}{Swin transformer: Hierarchical vision transformer using shifted windows}, in: \bibinfo{booktitle}{Proceedings of the IEEE/CVF international conference on computer vision}, pp. \bibinfo{pages}{10012--10022}.
\bibitem[{Luo et~al.(2022)Luo, Wang, Liao, Chen, Song, Chen, Zhang, Metaxas and Zhang}]{luo2022semi}
\bibinfo{author}{Luo, X.}, \bibinfo{author}{Wang, G.}, \bibinfo{author}{Liao, W.}, \bibinfo{author}{Chen, J.}, \bibinfo{author}{Song, T.}, \bibinfo{author}{Chen, Y.}, \bibinfo{author}{Zhang, S.}, \bibinfo{author}{Metaxas, D.N.}, \bibinfo{author}{Zhang, S.}, \bibinfo{year}{2022}.
\newblock \bibinfo{title}{Semi-supervised medical image segmentation via uncertainty rectified pyramid consistency}.
\newblock \bibinfo{journal}{Medical Image Analysis} \bibinfo{volume}{80}, \bibinfo{pages}{102517}.
\bibitem[{Maier-Hein et~al.(2017)Maier-Hein, Vedula, Speidel, Navab, Kikinis, Park, Eisenmann, Feussner, Forestier, Giannarou et~al.}]{maier2017surgical}
\bibinfo{author}{Maier-Hein, L.}, \bibinfo{author}{Vedula, S.S.}, \bibinfo{author}{Speidel, S.}, \bibinfo{author}{Navab, N.}, \bibinfo{author}{Kikinis, R.}, \bibinfo{author}{Park, A.}, \bibinfo{author}{Eisenmann, M.}, \bibinfo{author}{Feussner, H.}, \bibinfo{author}{Forestier, G.}, \bibinfo{author}{Giannarou, S.}, et~al., \bibinfo{year}{2017}.
\newblock \bibinfo{title}{Surgical data science for next-generation interventions}.
\newblock \bibinfo{journal}{Nature Biomedical Engineering} \bibinfo{volume}{1}, \bibinfo{pages}{691--696}.
\bibitem[{Morris(2005)}]{morris2005robotic}
\bibinfo{author}{Morris, B.}, \bibinfo{year}{2005}.
\newblock \bibinfo{title}{Robotic surgery: applications, limitations, and impact on surgical education}.
\newblock \bibinfo{journal}{Medscape General Medicine} \bibinfo{volume}{7}, \bibinfo{pages}{72}.
\bibitem[{Nishi et~al.(2024)Nishi, Kim, Li and Pfister}]{nishi2024joint}
\bibinfo{author}{Nishi, K.}, \bibinfo{author}{Kim, J.}, \bibinfo{author}{Li, W.}, \bibinfo{author}{Pfister, H.}, \bibinfo{year}{2024}.
\newblock \bibinfo{title}{Joint-task regularization for partially labeled multi-task learning}, in: \bibinfo{booktitle}{Proceedings of the IEEE/CVF Conference on Computer Vision and Pattern Recognition}, pp. \bibinfo{pages}{16152--16162}.
\bibitem[{Nwoye et~al.(2020)Nwoye, Gonzalez, Yu, Mascagni, Mutter, Marescaux and Padoy}]{nwoye2020recognition}
\bibinfo{author}{Nwoye, C.I.}, \bibinfo{author}{Gonzalez, C.}, \bibinfo{author}{Yu, T.}, \bibinfo{author}{Mascagni, P.}, \bibinfo{author}{Mutter, D.}, \bibinfo{author}{Marescaux, J.}, \bibinfo{author}{Padoy, N.}, \bibinfo{year}{2020}.
\newblock \bibinfo{title}{Recognition of instrument-tissue interactions in endoscopic videos via action triplets}, in: \bibinfo{booktitle}{International conference on medical image computing and computer-assisted intervention}, \bibinfo{organization}{Springer}. pp. \bibinfo{pages}{364--374}.
\bibitem[{Oh et~al.(2019)Oh, Lee, Xu and Kim}]{stm}
\bibinfo{author}{Oh, S.W.}, \bibinfo{author}{Lee, J.Y.}, \bibinfo{author}{Xu, N.}, \bibinfo{author}{Kim, S.J.}, \bibinfo{year}{2019}.
\newblock \bibinfo{title}{Video object segmentation using space-time memory networks}, in: \bibinfo{booktitle}{Proceedings of the IEEE/CVF international conference on computer vision}, pp. \bibinfo{pages}{9226--9235}.
\bibitem[{Pont-Tuset et~al.(2017)Pont-Tuset, Perazzi, Caelles, Arbel{\'a}ez, Sorkine-Hornung and Van~Gool}]{pont20172017}
\bibinfo{author}{Pont-Tuset, J.}, \bibinfo{author}{Perazzi, F.}, \bibinfo{author}{Caelles, S.}, \bibinfo{author}{Arbel{\'a}ez, P.}, \bibinfo{author}{Sorkine-Hornung, A.}, \bibinfo{author}{Van~Gool, L.}, \bibinfo{year}{2017}.
\newblock \bibinfo{title}{The 2017 davis challenge on video object segmentation}.
\newblock \bibinfo{journal}{arXiv preprint arXiv:1704.00675} .
\bibitem[{Qian et~al.(2020)Qian, Nan, Ancha, Okorn and Held}]{qian2020robust}
\bibinfo{author}{Qian, J.}, \bibinfo{author}{Nan, J.}, \bibinfo{author}{Ancha, S.}, \bibinfo{author}{Okorn, B.}, \bibinfo{author}{Held, D.}, \bibinfo{year}{2020}.
\newblock \bibinfo{title}{Robust instance tracking via uncertainty flow}.
\newblock \bibinfo{journal}{arXiv preprint arXiv:2010.04367} .
\bibitem[{Ramesh et~al.(2023)Ramesh, Dall'Alba, Gonzalez, Yu, Mascagni, Mutter, Marescaux, Fiorini and Padoy}]{ramesh2023weakly}
\bibinfo{author}{Ramesh, S.}, \bibinfo{author}{Dall'Alba, D.}, \bibinfo{author}{Gonzalez, C.}, \bibinfo{author}{Yu, T.}, \bibinfo{author}{Mascagni, P.}, \bibinfo{author}{Mutter, D.}, \bibinfo{author}{Marescaux, J.}, \bibinfo{author}{Fiorini, P.}, \bibinfo{author}{Padoy, N.}, \bibinfo{year}{2023}.
\newblock \bibinfo{title}{Weakly supervised temporal convolutional networks for fine-grained surgical activity recognition}.
\newblock \bibinfo{journal}{IEEE Transactions on Medical Imaging} \bibinfo{volume}{42}, \bibinfo{pages}{2592--2602}.
\bibitem[{Ravi et~al.(2025)Ravi, Gabeur, Hu, Hu, Ryali, Ma, Khedr, R{\"a}dle, Rolland, Gustafson et~al.}]{ravi2025sam}
\bibinfo{author}{Ravi, N.}, \bibinfo{author}{Gabeur, V.}, \bibinfo{author}{Hu, Y.T.}, \bibinfo{author}{Hu, R.}, \bibinfo{author}{Ryali, C.}, \bibinfo{author}{Ma, T.}, \bibinfo{author}{Khedr, H.}, \bibinfo{author}{R{\"a}dle, R.}, \bibinfo{author}{Rolland, C.}, \bibinfo{author}{Gustafson, L.}, et~al., \bibinfo{year}{2025}.
\newblock \bibinfo{title}{Sam 2: Segment anything in images and videos}, in: \bibinfo{booktitle}{International Conference on Learning Representations}, pp. \bibinfo{pages}{28085--28128}.
\bibitem[{Ren et~al.(2020)Ren, Yan, Yang, Yuille and Zha}]{ren2020unsupervised}
\bibinfo{author}{Ren, Z.}, \bibinfo{author}{Yan, J.}, \bibinfo{author}{Yang, X.}, \bibinfo{author}{Yuille, A.}, \bibinfo{author}{Zha, H.}, \bibinfo{year}{2020}.
\newblock \bibinfo{title}{Unsupervised learning of optical flow with patch consistency and occlusion estimation}.
\newblock \bibinfo{journal}{Pattern Recognition} \bibinfo{volume}{103}, \bibinfo{pages}{107191}.
\bibitem[{Rivoir et~al.(2020)Rivoir, Bodenstedt, Funke, von Bechtolsheim, Distler, Weitz and Speidel}]{rivoir2020rethinking}
\bibinfo{author}{Rivoir, D.}, \bibinfo{author}{Bodenstedt, S.}, \bibinfo{author}{Funke, I.}, \bibinfo{author}{von Bechtolsheim, F.}, \bibinfo{author}{Distler, M.}, \bibinfo{author}{Weitz, J.}, \bibinfo{author}{Speidel, S.}, \bibinfo{year}{2020}.
\newblock \bibinfo{title}{Rethinking anticipation tasks: Uncertainty-aware anticipation of sparse surgical instrument usage for context-aware assistance}, in: \bibinfo{booktitle}{International conference on medical image computing and computer-assisted intervention}, \bibinfo{organization}{Springer}. pp. \bibinfo{pages}{752--762}.
\bibitem[{Ronneberger et~al.(2015)Ronneberger, Fischer and Brox}]{unet}
\bibinfo{author}{Ronneberger, O.}, \bibinfo{author}{Fischer, P.}, \bibinfo{author}{Brox, T.}, \bibinfo{year}{2015}.
\newblock \bibinfo{title}{U-net: Convolutional networks for biomedical image segmentation}, in: \bibinfo{booktitle}{International Conference on Medical image computing and computer-assisted intervention}, \bibinfo{organization}{Springer}. pp. \bibinfo{pages}{234--241}.
\bibitem[{Ruder(2017)}]{ruder2017overview}
\bibinfo{author}{Ruder, S.}, \bibinfo{year}{2017}.
\newblock \bibinfo{title}{An overview of multi-task learning in deep neural networks}.
\newblock \bibinfo{journal}{arXiv preprint arXiv:1706.05098} .
\bibitem[{Rukundo(2024)}]{rukundo2024evaluation}
\bibinfo{author}{Rukundo, O.}, \bibinfo{year}{2024}.
\newblock \bibinfo{title}{Evaluation of extra pixel interpolation with mask processing for medical image segmentation with deep learning}.
\newblock \bibinfo{journal}{Signal, Image and Video Processing} \bibinfo{volume}{18}, \bibinfo{pages}{7703--7710}.
\bibitem[{Schreuder and Verheijen(2009)}]{schreuder2009robotic}
\bibinfo{author}{Schreuder, H.}, \bibinfo{author}{Verheijen, R.}, \bibinfo{year}{2009}.
\newblock \bibinfo{title}{Robotic surgery}.
\newblock \bibinfo{journal}{BJOG: An International Journal of Obstetrics \& Gynaecology} \bibinfo{volume}{116}, \bibinfo{pages}{198--213}.
\bibitem[{Sestini et~al.(2023)Sestini, Rosa, De~Momi, Ferrigno and Padoy}]{sestini2023fun}
\bibinfo{author}{Sestini, L.}, \bibinfo{author}{Rosa, B.}, \bibinfo{author}{De~Momi, E.}, \bibinfo{author}{Ferrigno, G.}, \bibinfo{author}{Padoy, N.}, \bibinfo{year}{2023}.
\newblock \bibinfo{title}{Fun-sis: A fully unsupervised approach for surgical instrument segmentation}.
\newblock \bibinfo{journal}{Medical Image Analysis} \bibinfo{volume}{85}, \bibinfo{pages}{102751}.
\bibitem[{Shi et~al.(2015)Shi, Yan, Xu and Jia}]{HMMN}
\bibinfo{author}{Shi, J.}, \bibinfo{author}{Yan, Q.}, \bibinfo{author}{Xu, L.}, \bibinfo{author}{Jia, J.}, \bibinfo{year}{2015}.
\newblock \bibinfo{title}{Hierarchical image saliency detection on extended cssd}.
\newblock \bibinfo{journal}{IEEE transactions on pattern analysis and machine intelligence} \bibinfo{volume}{38}, \bibinfo{pages}{717--729}.
\bibitem[{Shi et~al.(2021)Shi, Jin, Dou and Heng}]{shi2021semi}
\bibinfo{author}{Shi, X.}, \bibinfo{author}{Jin, Y.}, \bibinfo{author}{Dou, Q.}, \bibinfo{author}{Heng, P.A.}, \bibinfo{year}{2021}.
\newblock \bibinfo{title}{Semi-supervised learning with progressive unlabeled data excavation for label-efficient surgical workflow recognition}.
\newblock \bibinfo{journal}{Medical Image Analysis} \bibinfo{volume}{73}, \bibinfo{pages}{102158}.
\bibitem[{Shvets et~al.(2018)Shvets, Rakhlin, Kalinin and Iglovikov}]{shvets2018automatic}
\bibinfo{author}{Shvets, A.A.}, \bibinfo{author}{Rakhlin, A.}, \bibinfo{author}{Kalinin, A.A.}, \bibinfo{author}{Iglovikov, V.I.}, \bibinfo{year}{2018}.
\newblock \bibinfo{title}{Automatic instrument segmentation in robot-assisted surgery using deep learning}, in: \bibinfo{booktitle}{2018 17th IEEE international conference on machine learning and applications (ICMLA)}, \bibinfo{organization}{IEEE}. pp. \bibinfo{pages}{624--628}.
\bibitem[{Tarvainen and Valpola(2017)}]{tarvainen2017mean}
\bibinfo{author}{Tarvainen, A.}, \bibinfo{author}{Valpola, H.}, \bibinfo{year}{2017}.
\newblock \bibinfo{title}{Mean teachers are better role models: Weight-averaged consistency targets improve semi-supervised deep learning results}.
\newblock \bibinfo{journal}{Advances in neural information processing systems} \bibinfo{volume}{30}.
\bibitem[{Teed and Deng(2020)}]{teed2020raft}
\bibinfo{author}{Teed, Z.}, \bibinfo{author}{Deng, J.}, \bibinfo{year}{2020}.
\newblock \bibinfo{title}{Raft: Recurrent all-pairs field transforms for optical flow}, in: \bibinfo{booktitle}{European conference on computer vision}, \bibinfo{organization}{Springer}. pp. \bibinfo{pages}{402--419}.
\bibitem[{Thrun(1995)}]{thrun1995learning}
\bibinfo{author}{Thrun, S.}, \bibinfo{year}{1995}.
\newblock \bibinfo{title}{Is learning the n-th thing any easier than learning the first?}
\newblock \bibinfo{journal}{Advances in neural information processing systems} \bibinfo{volume}{8}.
\bibitem[{Twinanda et~al.(2016)Twinanda, Shehata, Mutter, Marescaux, De~Mathelin and Padoy}]{twinanda2016endonet}
\bibinfo{author}{Twinanda, A.P.}, \bibinfo{author}{Shehata, S.}, \bibinfo{author}{Mutter, D.}, \bibinfo{author}{Marescaux, J.}, \bibinfo{author}{De~Mathelin, M.}, \bibinfo{author}{Padoy, N.}, \bibinfo{year}{2016}.
\newblock \bibinfo{title}{Endonet: A deep architecture for recognition tasks on laparoscopic videos}.
\newblock \bibinfo{journal}{IEEE transactions on medical imaging} \bibinfo{volume}{36}, \bibinfo{pages}{86--97}.
\bibitem[{Valderrama et~al.(2022)Valderrama, Ruiz~Puentes, Hern{\'a}ndez, Ayobi, Verlyck, Santander, Caicedo, Fern{\'a}ndez and Arbel{\'a}ez}]{valderrama2022towards}
\bibinfo{author}{Valderrama, N.}, \bibinfo{author}{Ruiz~Puentes, P.}, \bibinfo{author}{Hern{\'a}ndez, I.}, \bibinfo{author}{Ayobi, N.}, \bibinfo{author}{Verlyck, M.}, \bibinfo{author}{Santander, J.}, \bibinfo{author}{Caicedo, J.}, \bibinfo{author}{Fern{\'a}ndez, N.}, \bibinfo{author}{Arbel{\'a}ez, P.}, \bibinfo{year}{2022}.
\newblock \bibinfo{title}{Towards holistic surgical scene understanding}, in: \bibinfo{booktitle}{International conference on medical image computing and computer-assisted intervention}, \bibinfo{organization}{Springer}. pp. \bibinfo{pages}{442--452}.
\bibitem[{Wang et~al.(2023)Wang, Chen, Wu, Luo, Tang, Dai, Zhao, Xie, Yuan and Jiang}]{ISVOS}
\bibinfo{author}{Wang, J.}, \bibinfo{author}{Chen, D.}, \bibinfo{author}{Wu, Z.}, \bibinfo{author}{Luo, C.}, \bibinfo{author}{Tang, C.}, \bibinfo{author}{Dai, X.}, \bibinfo{author}{Zhao, Y.}, \bibinfo{author}{Xie, Y.}, \bibinfo{author}{Yuan, L.}, \bibinfo{author}{Jiang, Y.G.}, \bibinfo{year}{2023}.
\newblock \bibinfo{title}{Look before you match: Instance understanding matters in video object segmentation}, in: \bibinfo{booktitle}{Proceedings of the IEEE/CVF conference on computer vision and pattern recognition}, pp. \bibinfo{pages}{2268--2278}.
\bibitem[{Wang et~al.(2022)Wang, Lu, Long, Zhong, Cheung, Dou and Liu}]{wang2022autolaparo}
\bibinfo{author}{Wang, Z.}, \bibinfo{author}{Lu, B.}, \bibinfo{author}{Long, Y.}, \bibinfo{author}{Zhong, F.}, \bibinfo{author}{Cheung, T.H.}, \bibinfo{author}{Dou, Q.}, \bibinfo{author}{Liu, Y.}, \bibinfo{year}{2022}.
\newblock \bibinfo{title}{Autolaparo: A new dataset of integrated multi-tasks for image-guided surgical automation in laparoscopic hysterectomy}, in: \bibinfo{booktitle}{Proceedings of the International Conference on Medical Image Computing and Computer-Assisted Intervention}, \bibinfo{organization}{Springer}. pp. \bibinfo{pages}{486--496}.
\bibitem[{Wei et~al.(2025)Wei, Budd, Garcia-Peraza-Herrera, Dorent, Shi and Vercauteren}]{wei2025segmatch}
\bibinfo{author}{Wei, M.}, \bibinfo{author}{Budd, C.}, \bibinfo{author}{Garcia-Peraza-Herrera, L.C.}, \bibinfo{author}{Dorent, R.}, \bibinfo{author}{Shi, M.}, \bibinfo{author}{Vercauteren, T.}, \bibinfo{year}{2025}.
\newblock \bibinfo{title}{Segmatch: semi-supervised surgical instrument segmentation}.
\newblock \bibinfo{journal}{Scientific Reports} \bibinfo{volume}{15}, \bibinfo{pages}{14042}.
\bibitem[{Wu et~al.(2023)Wu, Liu, Luo, Liu, Zheng, Liu, Jiang, Zhai and Wang}]{wu2023accflow}
\bibinfo{author}{Wu, G.}, \bibinfo{author}{Liu, X.}, \bibinfo{author}{Luo, K.}, \bibinfo{author}{Liu, X.}, \bibinfo{author}{Zheng, Q.}, \bibinfo{author}{Liu, S.}, \bibinfo{author}{Jiang, X.}, \bibinfo{author}{Zhai, G.}, \bibinfo{author}{Wang, W.}, \bibinfo{year}{2023}.
\newblock \bibinfo{title}{Accflow: Backward accumulation for long-range optical flow}, in: \bibinfo{booktitle}{Proceedings of the IEEE/CVF International Conference on Computer Vision}, pp. \bibinfo{pages}{12119--12128}.
\bibitem[{Xie et~al.(2024)Xie, Xie and Zisserman}]{xie2024appearance}
\bibinfo{author}{Xie, J.}, \bibinfo{author}{Xie, W.}, \bibinfo{author}{Zisserman, A.}, \bibinfo{year}{2024}.
\newblock \bibinfo{title}{Appearance-based refinement for object-centric motion segmentation}, in: \bibinfo{booktitle}{European Conference on Computer Vision}, \bibinfo{organization}{Springer}. pp. \bibinfo{pages}{238--256}.
\bibitem[{Xu et~al.(2022)Xu, Zhang, Cai, Rezatofighi and Tao}]{gmflow}
\bibinfo{author}{Xu, H.}, \bibinfo{author}{Zhang, J.}, \bibinfo{author}{Cai, J.}, \bibinfo{author}{Rezatofighi, H.}, \bibinfo{author}{Tao, D.}, \bibinfo{year}{2022}.
\newblock \bibinfo{title}{Gmflow: Learning optical flow via global matching}, in: \bibinfo{booktitle}{Proceedings of the IEEE/CVF conference on computer vision and pattern recognition}, pp. \bibinfo{pages}{8121--8130}.
\bibitem[{Yi and Jiang(2019)}]{yi2019hard}
\bibinfo{author}{Yi, F.}, \bibinfo{author}{Jiang, T.}, \bibinfo{year}{2019}.
\newblock \bibinfo{title}{Hard frame detection and online mapping for surgical phase recognition}, in: \bibinfo{booktitle}{International Conference on Medical Image Computing and Computer-Assisted Intervention}, \bibinfo{organization}{Springer}. pp. \bibinfo{pages}{449--457}.
\bibitem[{Yin et~al.(2025)Yin, Wang, Ye, Meng and Fu}]{yin2025memory}
\bibinfo{author}{Yin, M.}, \bibinfo{author}{Wang, F.}, \bibinfo{author}{Ye, X.}, \bibinfo{author}{Meng, Y.}, \bibinfo{author}{Fu, Z.}, \bibinfo{year}{2025}.
\newblock \bibinfo{title}{Memory-augmented sam2 for training-free surgical video segmentation}, in: \bibinfo{booktitle}{International Conference on Medical Image Computing and Computer-Assisted Intervention}, \bibinfo{organization}{Springer}. pp. \bibinfo{pages}{328--337}.
\bibitem[{Yu et~al.(2025)Yu, Wang, Dong, Xu, Islam, Wang, Bai and Ren}]{yu2025sam}
\bibinfo{author}{Yu, J.}, \bibinfo{author}{Wang, A.}, \bibinfo{author}{Dong, W.}, \bibinfo{author}{Xu, M.}, \bibinfo{author}{Islam, M.}, \bibinfo{author}{Wang, J.}, \bibinfo{author}{Bai, L.}, \bibinfo{author}{Ren, H.}, \bibinfo{year}{2025}.
\newblock \bibinfo{title}{Sam 2 in robotic surgery: An empirical evaluation for robustness and generalization in surgical video segmentation}, in: \bibinfo{booktitle}{International Workshop on Efficient Medical Artificial Intelligence}, \bibinfo{organization}{Springer}. pp. \bibinfo{pages}{174--183}.
\bibitem[{Yu et~al.(2022)Yu, Zhao, Jin, Chen, Dou and Heng}]{yu2022pseudo}
\bibinfo{author}{Yu, Y.}, \bibinfo{author}{Zhao, Z.}, \bibinfo{author}{Jin, Y.}, \bibinfo{author}{Chen, G.}, \bibinfo{author}{Dou, Q.}, \bibinfo{author}{Heng, P.A.}, \bibinfo{year}{2022}.
\newblock \bibinfo{title}{Pseudo-label guided cross-video pixel contrast for robotic surgical scene segmentation with limited annotations}, in: \bibinfo{booktitle}{2022 IEEE/RSJ International Conference on Intelligent Robots and Systems (IROS)}, \bibinfo{organization}{IEEE}. pp. \bibinfo{pages}{10857--10864}.
\bibitem[{Zhang et~al.(2025)Zhang, Lv, Xue, Zhang, Liu, Fu, Cheng and Qi}]{zhang2025semisam+}
\bibinfo{author}{Zhang, Y.}, \bibinfo{author}{Lv, B.}, \bibinfo{author}{Xue, L.}, \bibinfo{author}{Zhang, W.}, \bibinfo{author}{Liu, Y.}, \bibinfo{author}{Fu, Y.}, \bibinfo{author}{Cheng, Y.}, \bibinfo{author}{Qi, Y.}, \bibinfo{year}{2025}.
\newblock \bibinfo{title}{Semisam+: rethinking semi-supervised medical image segmentation in the era of foundation models}.
\newblock \bibinfo{journal}{Medical Image Analysis} , \bibinfo{pages}{103733}.
\bibitem[{Zhang et~al.(2024)Zhang, Wang, Pan, Jiang, Ge, Guo, Jiang, Lu, Zhang, Liu et~al.}]{zhang2024nasalseg}
\bibinfo{author}{Zhang, Y.}, \bibinfo{author}{Wang, J.}, \bibinfo{author}{Pan, T.}, \bibinfo{author}{Jiang, Q.}, \bibinfo{author}{Ge, J.}, \bibinfo{author}{Guo, X.}, \bibinfo{author}{Jiang, C.}, \bibinfo{author}{Lu, J.}, \bibinfo{author}{Zhang, J.}, \bibinfo{author}{Liu, X.}, et~al., \bibinfo{year}{2024}.
\newblock \bibinfo{title}{Nasalseg: A dataset for automatic segmentation of nasal cavity and paranasal sinuses from 3d ct images}.
\newblock \bibinfo{journal}{Scientific Data} \bibinfo{volume}{11}, \bibinfo{pages}{1329}.
\bibitem[{Zhang and Yang(2021)}]{zhang2021survey}
\bibinfo{author}{Zhang, Y.}, \bibinfo{author}{Yang, Q.}, \bibinfo{year}{2021}.
\newblock \bibinfo{title}{A survey on multi-task learning}.
\newblock \bibinfo{journal}{IEEE transactions on knowledge and data engineering} \bibinfo{volume}{34}, \bibinfo{pages}{5586--5609}.
\bibitem[{Zhao et~al.(2020)Zhao, Jin, Gao, Dou and Heng}]{zhao2020learning}
\bibinfo{author}{Zhao, Z.}, \bibinfo{author}{Jin, Y.}, \bibinfo{author}{Gao, X.}, \bibinfo{author}{Dou, Q.}, \bibinfo{author}{Heng, P.A.}, \bibinfo{year}{2020}.
\newblock \bibinfo{title}{Learning motion flows for semi-supervised instrument segmentation from robotic surgical video}, in: \bibinfo{booktitle}{International Conference on Medical Image Computing and Computer-Assisted Intervention}, \bibinfo{organization}{Springer}. pp. \bibinfo{pages}{679--689}.

\end{thebibliography}







\end{document}